\documentclass[acmlarge]{acmart}
\AtBeginDocument{%
  \providecommand\BibTeX{{%
    \normalfont B\kern-0.5em{\scshape i\kern-0.25em b}\kern-0.8em\TeX}}}

\setcopyright{acmcopyright}
\copyrightyear{2018}
\acmYear{2018}
\acmDOI{XXXXXXX.XXXXXXX}

\acmJournal{POMACS}
\acmVolume{37}
\acmNumber{4}
\acmArticle{111}
\acmMonth{8}

\usepackage{hyperref}
\usepackage{mathrsfs}

\usepackage{amsfonts,amssymb}
\usepackage{multirow}
\usepackage{booktabs}
\usepackage{subfigure}
\usepackage{bbding}
\usepackage{amsmath}
\usepackage{CJK}
\usepackage{indentfirst}
\usepackage{makecell}

\usepackage{tabularx}
\usepackage{wrapfig}

\begin{document}

% \title{Visual Adversarial Attacks and Defenses in the Physical World:\ \ \ \ \ \ \ \ \ \ \ \ \ \ \   A Survey}
\title{Visual Adversarial Attacks and Defenses in the Physical World: A Survey}
%% use to denote shared contribution to the research.

\author{Xingxing Wei}
\authornote{Xingxing Wei is the corresponding author}
\email{xxwei@buaa.edu.cn}
\author{Bangzheng Pu}
\email{Pu_bangzheng@buaa.edu.cn}
\author{Shiji Zhao}
\email{zhaoshiji123@buaa.edu.cn}
\author{Jiefan Lu}
\email{jf_lu@buaa.edu.cn}
\affiliation{%
  \institution{Institute of Artificial Intelligence, Beihang University}
  \streetaddress{No.37, Xueyuan Road}
  \city{Beijing}
  \country{P.R. China}
  \postcode{100191}
}

\author{Baoyuan Wu}
\email{wubaoyuan@cuhk.edu.cn}
\affiliation{
    \institution{The Chinese University of HongKong}
    \city{Shenzhen}
    \country{P.R.China}
}

\renewcommand{\shortauthors}{Xingxing Wei, Bangzheng Pu, Shiji Zhao, Jiefan Lu, Baoyuan Wu.}

\begin{abstract}
Although Deep Neural Networks (DNNs) have been widely applied in various real-world scenarios, they remain vulnerable to adversarial examples. Adversarial attacks in computer vision can be categorized into digital attacks and physical attacks based on their different forms. Compared to digital attacks, which generate perturbations in digital pixels, physical attacks are more practical in real-world settings. Due to the serious security risks posed by physically adversarial examples, many studies have been conducted to evaluate the physically adversarial robustness of DNNs in recent years. In this paper, we provide a comprehensive survey of current physically adversarial attacks and defenses in computer vision. We establish a taxonomy by organizing physical attacks according to attack tasks, attack forms, and attack methods. This approach offers readers a systematic understanding of the topic from multiple perspectives. For physical defenses, we categorize them into pre-processing, in-processing, and post-processing for DNN models to ensure comprehensive coverage of adversarial defenses. Based on this survey, we discuss the challenges facing this research field and provide an outlook on future directions.
\end{abstract}





\maketitle

\section{\textbf{Introduction}}
Deep Neural Networks (DNNs) have achieved impressive performance in Computer Vision (CV) and Natural Language Processing (NLP), and thus are widely applied in businesses and industries, such as mobile payment \cite{du2018mobile}, autonomous driving \cite{badue2021self}, surveillance \cite{ansari2021human}, robotics \cite{polydoros2017survey}, and medical diagnosis \cite{bhattacharya2021deep}, etc. However, the potential crises hide behind the prosperity.  Szegedy et al. \cite{goodfellow2014explaining} discover that only a tiny perturbation can overthrow the correct predictions of state-of-the-art DNN models. This malicious behavior is defined as the Adversarial Attack and the manipulated image is called the Adversarial Example. After that, the effectiveness of adversarial examples has been proved in different CV tasks, which mainly contain image classification \cite{madry2018towards}, object detection \cite{liang2021parallel}, object tracking \cite{yan2020hijacking} etc. Moreover, some research \cite{liu2017delving, ilyas2019adversarial} has also theoretically demonstrated the vulnerability of DNNs.

In computer vision, adversarial attacks can be categorized into digital attacks and physical attacks based on the domain of implementation. Digital attacks are carried out on digital pixels after camera imaging, while physical attacks target physical objects before camera imaging. Although digital attacks, such as PGD \cite{madry2018towards}, MI-FGSM \cite{dong2018boosting}, C\&W \cite{carlini2017towards}, and Deepfool \cite{moosavi2016deepfool}, have been effective against DNN models, they are challenging to implement in real-world scenarios. This difficulty arises because digital perturbations are often global and inconspicuous, making them hard to capture clearly by the camera. These limitations have driven researchers to explore more practical attack methods in the physical world. As a result, many physically adversarial attacks have been reported in various domains, such as autonomous driving \cite{lu2017adversarial, chen2018shapeshifter, kong2020physgan, bai2021inconspicuous, zhang2019camou, duan2020adversarial, xue2021naturalae}, face recognition \cite{sharif2016accessorize, lu2017adversarial, pautov2019adversarial, singh2021brightness, sharif2019general, xiao2021improving}, and security surveillance \cite{thys2019fooling, wu2020making, hu2022adversarial, wiyatno2019physical, ding2021towards, hu2021naturalistic}. These studies highlight significant challenges for safety-critical tasks, making it crucial to investigate the robustness of CV systems against physical attacks and to develop corresponding defense strategies.

Generally speaking, digital attacks and physical attacks occur at different stages within the visual recognition pipeline. For example, Fig.~\ref{attack_pipeline} illustrates the distinction between physical attacks and digital attacks through the traffic sign recognition task in a self-driving scenario. We can see that for a physical attack, adversaries can manipulate the real-world object (e.g., pasting an adversarial patch on the stop sign) or interfere with the camera's imaging process (e.g., placing transparent adversarial films or optical elements in front of the lens). In the case of a digital attack, adversaries typically generate adversarial perturbations directly on the pixels within the captured image or video. Regardless of the type of attack, when the generated adversarial examples are input into the DNN models, they lead to incorrect predictions. Compared to digital attacks, physical attacks face several crucial challenges:

\begin{figure*}[ht]
\centering
\includegraphics[width=1.0\textwidth]{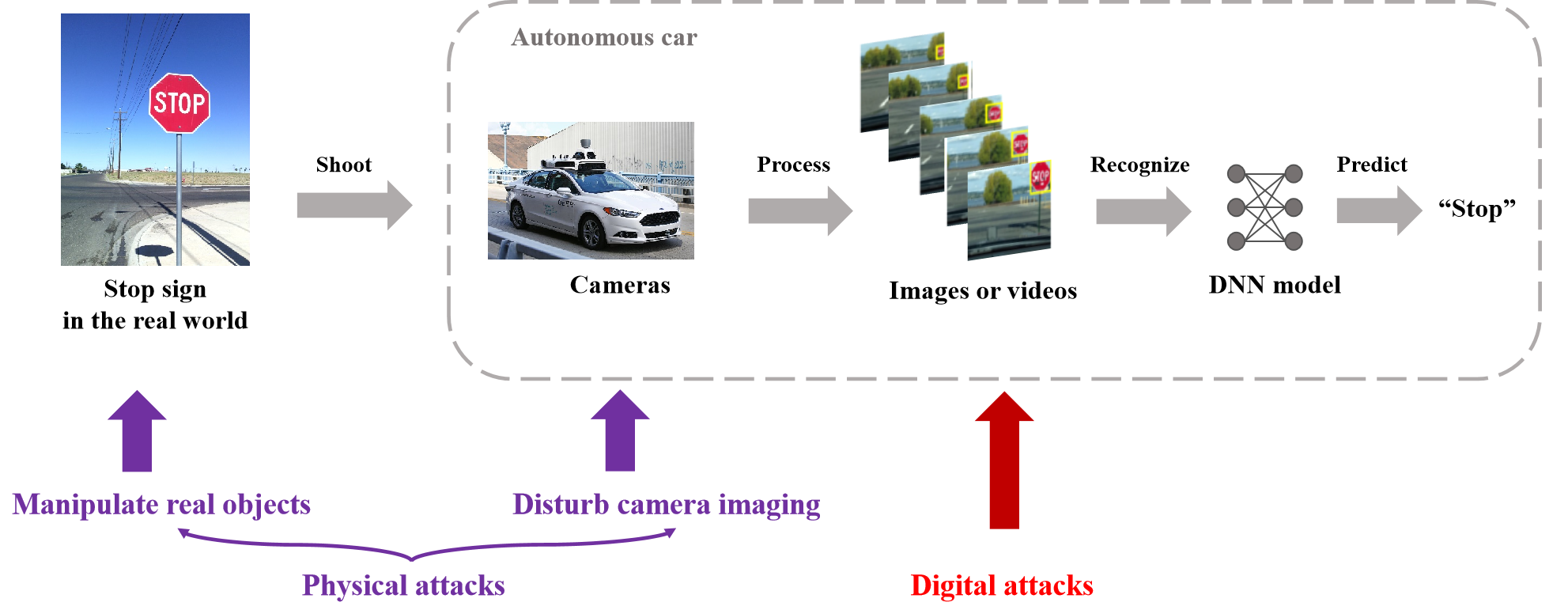}
\caption{The comparison between digital attacks and physical attacks in the standard visual recognition pipeline. Here we choose traffic sign recognition as the example to show their differences.}
\label{attack_pipeline}
\end{figure*}

\textcolor{black}{(1) When manipulating real objects, the physically adversarial example must withstand the effects of camera imaging and various transformations. Because camera imaging, including optical lens and image sensor processor, as well as various transformations, including changes in shooting distance, angle and illumination, will cause the appearance distortion and shape deformation of the carefully crafted adversarial examples, leading to a major distinction between the designed adversarial examples and captured adversarial examples, and thus will reduce their attack ability in practical applications.}

\textcolor{black}{(2) For disturbing camera imaging, a fundamental challenge lies in the inherent trade-off between attack effectiveness and imaging integrity. On one hand, the adversarial medium must be physically realizable and unobtrusive, ensuring that it does not noticeably degrade the image quality or interfere with the camera’s normal functioning. On the other hand, it must introduce sufficient perturbation in the captured images to fool deep learning models into making incorrect predictions. }

\textcolor{black}{(3) For both physical attacks above, they should be stealthy for human eyes. Digital attacks are modified at the pixel level. The generated adversarial perturbation is often constrained within a very small range, making it difficult to be detected by the human eye.  However, it is challenging for physical attacks to realize inconspicuousness as they are often designed to be visible in order to ensure they can be captured by cameras. }

\begin{figure*}[t]
\centering
\includegraphics[width=1\textwidth]{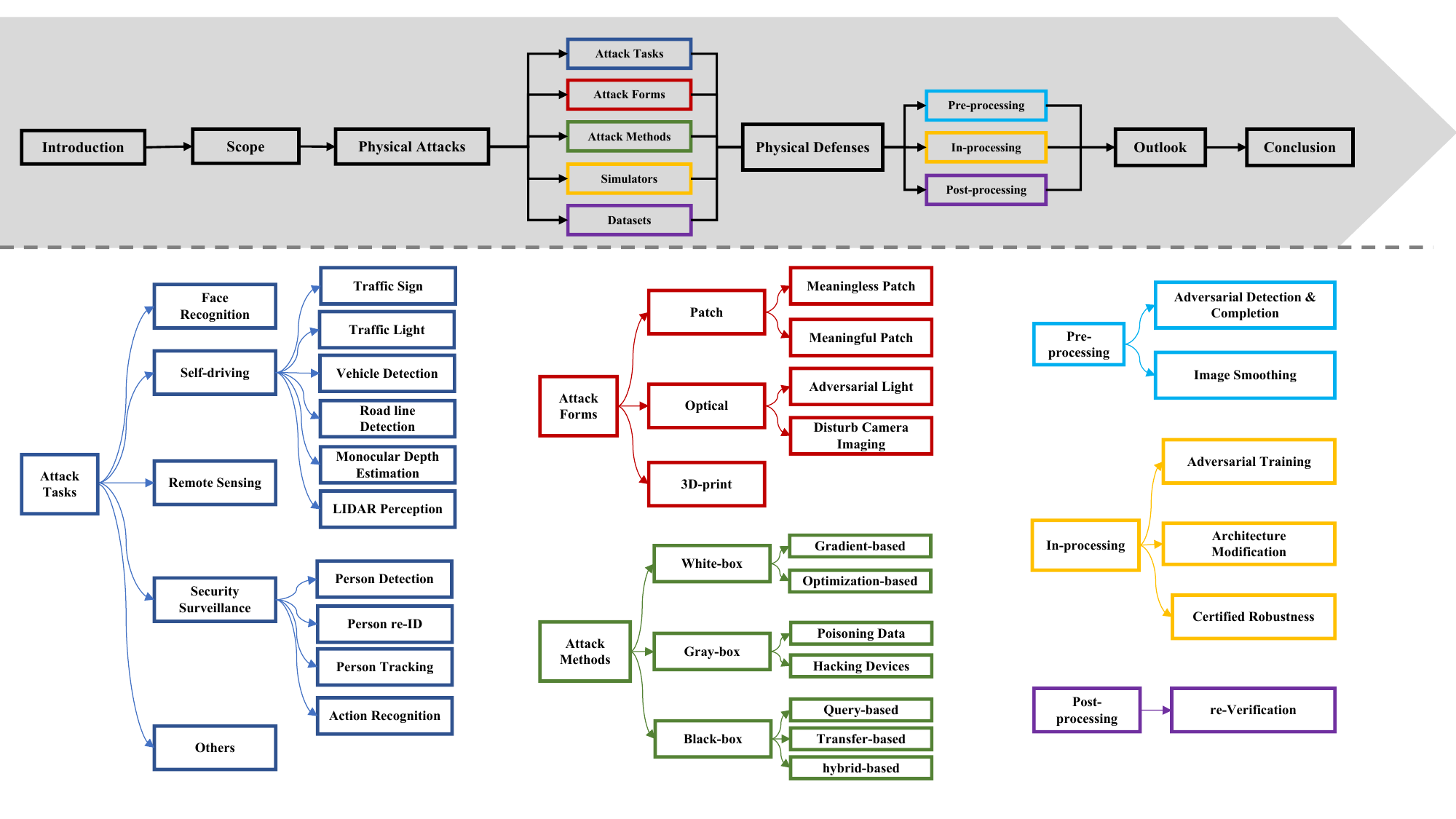} 
\caption{The framework of this survey. We first establish the taxonomy of physically adversarial attacks and defenses, and then, the outlook for the future direction is discussed.}
\label{survey}
\end{figure*}

\begin{wraptable}{r}{0.6\textwidth}
  \centering
  % \vspace{-18pt} % 适当调节垂直位置
  \caption{Published surveys of adversarial attacks and defenses in computer vision. The number of referred papers about physical attacks and defenses is shown in the last column.}
  % \vspace{5pt}
  \begin{tabular}{ccccc}
    \toprule
    Surveys  & \makecell[c]{Physical \\ attacks} & \makecell[c]{Physical \\ defenses} & Year & Number\\
    \midrule
    Chakraborty et al.\cite{chakraborty2018adversarial} & \XSolid & \XSolid & 2018 & 0 \\
    Akhtar et al.\cite{akhtar2018threat} & \Checkmark & \XSolid & 2018 & 8 \\
    Qiu et al.\cite{qiu2019review} & \Checkmark & \XSolid & 2019 & 8 \\
    Serban et al.\cite{serban2020adversarial} & \Checkmark & \XSolid & 2020 & 4 \\
    Huang et al.\cite{huang2020survey} & \Checkmark & \XSolid & 2020 & 1 \\
    Machado et al.\cite{machado2021adversarial} & \Checkmark & \XSolid & 2021 & 5 \\
    Akhtar et al.\cite{akhtar2021advances} & \Checkmark & \XSolid & 2021 & 34 \\
    Wei et al.\cite{wei2024physical} & \Checkmark & \XSolid & 2024 & 46 \\
    Ours & \Checkmark & \Checkmark & 2025 & \textbf{151} \\
    \bottomrule
  \end{tabular}
  \label{statistic}
\end{wraptable}
To address the above issues, numerous studies have been conducted over the past years to evaluate the physically adversarial robustness of DNNs. It is necessary to explore the current physical attack and defense methods, and to summarize the existing progress and future directions. Although there are several surveys on adversarial attacks and defenses in computer vision, they are neither comprehensive nor up-to-date regarding physical attacks and defenses. For example, \cite{chakraborty2018adversarial} discusses digital attacks and defenses focusing on various machine learning models, but does not cover physical attacks and defenses. Akhtar et al. \cite{akhtar2018threat} review adversarial attacks and defenses in computer vision up to 2018, including only a few early-stage physical attack methods under "Attacks in the real world." In 2019, Qiu et al. \cite{qiu2019review} introduced physically adversarial attacks as an application scenario of adversarial attacks. In 2020, Huang et al. \cite{huang2020survey} summarized studies in the field into five parts: verification, testing, adversarial attacks and defenses, and interpretability. Later, \cite{akhtar2021advances} systematically reviewed adversarial attacks and defenses, but physical attacks were still not the main focus. Other works have concentrated on specific computer vision tasks, such as Serban et al. \cite{serban2020adversarial}, who summarized adversarial attacks and defenses on object recognition, and Machado et al. \cite{machado2021adversarial}, who reviewed adversarial machine learning in image classification. These surveys primarily focus on digital domain adversarial attacks and defenses, largely neglecting physical attacks. Recently, Wei et al. \cite{wei2024physical} published a survey on physically adversarial attacks in computer vision, discussing current physical attacks in terms of effectiveness, stealthiness, and robustness. Our survey differs from theirs in three key aspects: (1) In addition to physically adversarial attacks, we also review physically adversarial defenses, which \cite{wei2024physical} does not cover. (2) We provide a more refined classification of current physically adversarial attacks from multiple perspectives (see Fig.~\ref{survey}), offering a more systematic approach than \cite{wei2024physical}. (3) We discuss a greater number of referenced papers on physical attacks and defenses than \cite{wei2024physical} (151 vs. 46). A comparison of these surveys is listed in Table \ref{statistic}. In brief, a comprehensive review of the latest physical attacks and defenses in computer vision is imperative.
 
 Therefore, this paper provides a survey to systematically record the recent research in this field over the last ten years. Specifically, we first establish a taxonomy of the current physical attacks from attack tasks, attack forms, and attack methods, respectively. For attack tasks, we introduce five safety-critical tasks where researchers usually conduct physical attacks. For attack forms, we conclude three main ways in which the current physical attacks are involved. For attack methods, we summarize four attack settings from the white-box and black-box perspectives. Secondly, we establish a taxonomy of the current physical defenses from three stages of a DNN model: pre-processing, in-processing, and post-processing. Each stage contains multiple fine-grained classifications for the defense methods. Finally, we discuss seven existing challenges that are still unsolved in this field and provide an outlook on future directions. The framework of this survey is illustrated in Fig~\ref{survey}.

In summary, this survey has three main contributions:

(1) We establish a taxonomy of the current physical attacks from three perspectives: attack tasks, attack forms, and attack methods. We believe these three aspects can give a clear description to readers.

(2) We establish a taxonomy of the current physical defenses from three perspectives: pre-processing, in-processing, and post-processing for the DNN models to achieve full coverage of the defense process. 

(3) Physical attacks and defenses are still a rapidly developing research field, and many problems need to be solved. Thus, we summarize challenges in this field and offer a prospect on the future direction. 

The rest of this paper is structured as follows:  In section \ref{section2}, we introduce the physically adversarial attacks. In section \ref{section3}, we review the physically adversarial defenses. In section \ref{section4}, we analyze challenges in this field and provide an outlook on this research direction.

\begin{table*}[htbp]
  \centering
  \caption{Search Terms for Literature Collection.}
  \begin{tabularx}{\textwidth}{lX}
    \toprule
    \textbf{Category} & \textbf{Search Terms} \\
    \midrule
    Primary Search Terms &
    Adversarial attacks, Physical adversarial attacks, adversarial defense, adversarial examples \\
    \midrule
    Attack Modalities &
    Adversarial Patch, adversarial texture, adversarial camouflage, adversarial light, mesh, 3D print \\
    \midrule
    Targeted Systems and Tasks & 
    Object detection, person detection, face recognition/detection, object detectors, person re-identification, self-driving, autonomous driving, 
    LIDAR perception, monocular depth estimation \\
    \midrule
    Defense Mechanisms and Robustness &
    Adversarial training, certified defense, adversarial distillation, randomized smoothing, adversarial patch detection, robustness \\
    \midrule
    Real-World Context &
    Physical world, real-world, stealthy/naturalness/imperceptible attack 
    \\
    \bottomrule
  \end{tabularx}
  \label{tab:searchterms}
\end{table*}

\section{\textbf{Scope of the Study}}
\begin{wrapfigure}{r}{0.45\textwidth}
  \centering
    \vspace{-10pt} % 适当调节垂直位置
  \includegraphics[width=0.45\textwidth]{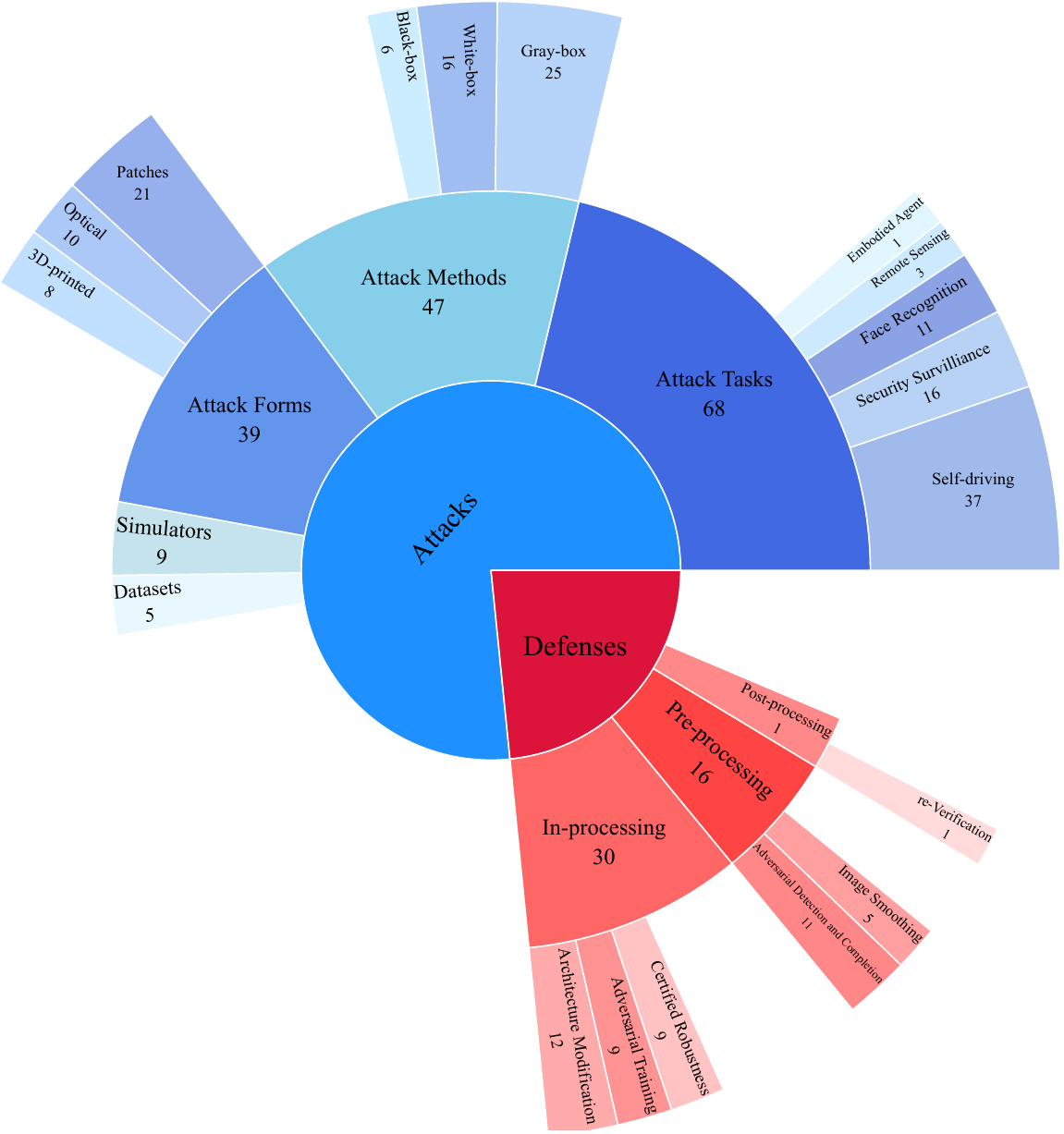}
  \caption{\textcolor{black}{Distribution of papers on physical adversarial attacks and defenses.}}
    \vspace{-10pt} % 适当调节垂直位置
  \label{numofpapers}
\end{wrapfigure}
\textcolor{black}{Our literature review methodology was designed to ensure a thorough and unbiased survey of physical-world adversarial attacks and defenses. We built a repository of over 200 papers, primarily from top-tier computer vision and security venues such as CVPR, ICCV, ECCV, NeurIPS, USENIX Security, AAAI, IEEE Transactions (e.g., T-PAMI, TIP), and ACM CSUR. To ensure relevance, papers had to meet strict criteria: directly address physical-world attacks or defenses in vision systems; involve core vision tasks (e.g., object detection, face recognition, depth estimation); use key physical sensors (e.g., RGB, infrared, LiDAR); and have notable impact (peer-reviewed or widely cited). We excluded work limited to digital attacks, unrelated to vision, duplicative, or in non-English. Based on this, we selected over 100 representative papers. Fig.~\ref{numofpapers} shows the distribution by category.}

\textcolor{black}{This survey identifies and analyzes current techniques for physical-world adversarial attacks and defenses. Our goals are: (1) to classify attack methods by physical medium (e.g., patches, 3D objects, wearables, light), target application (e.g., autonomous driving, surveillance), and attack type (white-box, black-box, gray-box); and (2) to review defense methods, covering passive strategies (e.g., patch detection), proactive hardening (e.g., adversarial training, knowledge distillation), and certified defenses (e.g., randomized smoothing). Table~\ref{tab:searchterms} summarizes the search parameters.}

\textcolor{black}{This study reviews literature from 2015–2025 to trace the evolution of physical adversarial research. Before 2016, work focused on deep learning theory and digital attacks, laying the foundation for physical methods. From 2016–2018, physical attacks gained attention, showing that digital perturbations could be transferred to the real world via printed images. This phase saw the first targeted attacks on face recognition and traffic signs. Between 2019–2025, attacks became more diverse and stealthy (e.g., using adversarial clothing, 3D objects, lasers, and light). Attacks on LiDAR and infrared also emerged. In parallel, defenses advanced from basic input transformations to robust methods, such as Vision Transformer-based defenses and diffusion model purification, highlighting the evolving arms race between attackers and defenders.}

\section{\textbf{Physically Adversarial Attacks}} 
\label{section2}
Physical attacks are carried out in the real world and are becoming more threatening than digital attacks. Since Kurakin et al. \cite{kurakin2016adversarial} first demonstrated that DNNs are vulnerable to physically adversarial examples, a lot of attention has been poured into the physical domain. In this section, we introduce different attack tasks in physical scenarios, discuss physical forms of adversarial examples, and analyze attack methods in the white-box setting and the black-box setting (Representative methods in the different attack tasks are briefly exhibited in Table \ref{FR}, \ref{SD}, \ref{SS}, respectively).

\begin{table*}[htbp]
  \centering
  \caption{Representative physical attacks against \textbf{face recognition systems.}}
    \begin{tabular}{ccccc}
    \toprule
    Task  & Methods & Settings & Physical forms & Sources \\
    \midrule
    \multicolumn{1}{c}{\multirow{9}[2]{*}{Face recognition}} & Adv-Glass \cite{sharif2016accessorize} & White\&Black & Meaningless Patch (Glassframe) &  SIGSAC 2016 \\
          & AGNs \cite{sharif2019general} & Black & Meaningful Patch (Glassframe) & TOPS 2019 \\
          & Pautov et al. \cite{pautov2019adversarial} & White & Meaningless Patch (Glassframe) & SIBIRCON 2019 \\
          & Adv-Hat \cite{komkov2021advhat} & White & Meaningless Patch (Hat) & ICPR 2020 \\
          & Nguyen et al. \cite{nguyen2020adversarial} & White & Adversarial Light & CVPR 2020 \\
          % & GenAP-DI \cite{xiao2021improving} & Black & Meaningful Patch (Makeup) & CVPR 2021 \\
          & Adv-Makeup \cite{yin2021adv} & Black & Meaningful Patch (Makeup) & IJCAI 2021 \\
          & Wei et al. \cite{wei2021generating} & Black & Meaningless Patch (Sticker) & TPAMI 2022 \\
          & Adv-Sticker \cite{wei2022adversarial} & Black & Meaningful Patch (Sticker) & TPAMI 2022 \\
    \bottomrule
    \end{tabular}%
  \label{FR}%
\end{table*}%

\subsection{\textbf{Attack tasks}}

Physical attacks are relevant in real-world scenarios that encompass a wide range of critical tasks. In this context, we primarily introduce several safety-critical applications, including face recognition, autonomous driving, security surveillance, and remote sensing. Additionally, we discuss the potential risks of physical attacks against embodied agents.

\subsubsection{\textbf{Face recognition}}

Concerns about face security in identity verification and related fields are increasing. Attackers use wearable physical adversarial examples to impersonate others, threatening privacy and property. These attacks are generally divided into two types: dodging and impersonation. Dodging attacks aims to misclassify a face as any other identity, minimizing recognition of the true identity. Impersonation attacks try to fool the system into recognizing a face as a specific target identity by minimizing the difference between the output and the target.

Early work focused on white-box attacks. Sharif et al. \cite{sharif2016accessorize} created adversarial glasses to evade or impersonate local face models. To improve transferability, patches of various shapes were designed to attack commercial models~\cite{komkov2021advhat,pautov2019adversarial}. Physical adversarial patches require robustness against environmental changes, but once printed, they are hard to update. However, adversarial light can be adapted more easily. Zhou et al. \cite{zhou2018invisible} used infrared LED hats to deceive face recognition without contact. Nguyen et al. \cite{nguyen2020adversarial} projected digital patches onto faces using standard projectors.

Black-box attacks are more practical in the physical world since many face recognition systems use commercial APIs that attackers can query. Sharif et al. \cite{sharif2016accessorize} updated patterns by querying target models. Finding vulnerable patch locations is important to improve effectiveness and reduce overfitting. Wei et al. \cite{wei2022adversarial} developed a query-based method to find optimal patch positions and rotations for cartoon stickers without calculating perturbations. To reduce query numbers, Guo et al. \cite{wei2021generating} jointly optimized patch location and perturbation, cutting queries from thousands to dozens.

\begin{figure}[ht]
    \centering
    \subfigure[Glass \cite{sharif2019general}]{
        \includegraphics[width=0.15\textwidth]{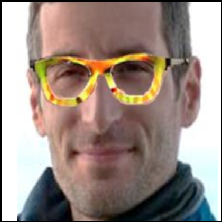}
    }
    \subfigure[Hat \cite{komkov2021advhat}]{
        \includegraphics[width=0.15\textwidth]{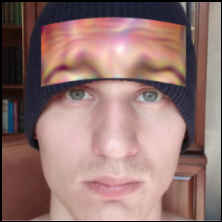}
    }
    \subfigure[Patch \cite{wei2021generating}]{
        \includegraphics[width=0.15\textwidth]{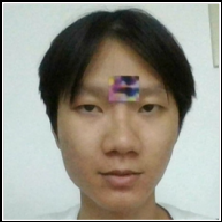}
    }
    \subfigure[Sticker \cite{wei2022adversarial}]{
        \includegraphics[width=0.15\textwidth]{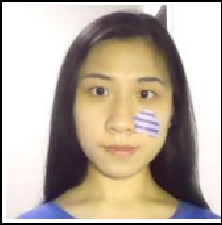}
    }
    \subfigure[Makeup \cite{yin2021adv}]{
        \includegraphics[width=0.15\textwidth]{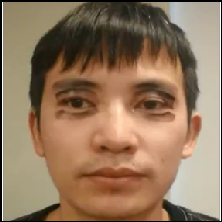}
    }
    \subfigure[Projection \cite{nguyen2020adversarial}]{
        \includegraphics[width=0.15\textwidth]{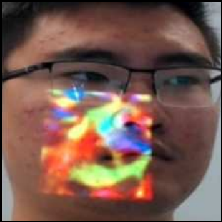}
    }
    \caption{Representative Methods of physically adversarial attacks against face recognition.}
    \label{face_recognition}
\end{figure}

% &  Lu et al. \cite{lu2017adversarial} & White & Meaningless Patch & ArXiv 2017 \\
\begin{table*}[htbp]
  \centering
  \caption{Representative physical attacks against \textbf{autonomous driving systems}. UW: USENIX WOOT, US: USENIX Security, IA: IEEE Access.}
    % \begin{tabular}{ccccc}
    \begin{tabular}{p{1.2cm}cccc}
    \midrule
    \multicolumn{1}{c}{Sub-tasks} & Methods & Settings & Physical Forms & \multicolumn{1}{c}{Sources} \\
    \midrule
    \multicolumn{1}{c}{\multirow{9}[2]{*}{Traffic sign recognition}} & RP2 \cite{eykholt2018robust} & White & Meaningless Patch & CVPR 2018 \\
    \multicolumn{1}{c}{} & RP2D \cite{song2018physical} & White & Meaningless Patch & UW 2018 \\
    \multicolumn{1}{c}{} & DT-UAPs \cite{benz2020double} & White & Meaningless Patch & ACCV 2020 \\
    \multicolumn{1}{c}{} & AdvCam \cite{duan2020adversarial} & White & Meaningful Patch & CVPR 2020 \\
    \multicolumn{1}{c}{} & IAP \cite{bai2021inconspicuous} & Black & Meaningless Patch & IOJT 2021 \\
    \multicolumn{1}{c}{} & SLAP \cite{lovisotto2021slap} & White & Adversarial Light  & US 2021 \\
    \multicolumn{1}{c}{} & Adv-Shadow \cite{zhong2022shadows} &Black  & Meaningful Patch & CVPR 2022 \\
    \multicolumn{1}{c}{} & RFLA \cite{wang2023rfla} &Black  & Adversarial Light & CVPR 2023 \\
    \multicolumn{1}{c}{} & ILRs\cite{sato2024invisible} &Black  & Adversarial Light & NDSS 2024 \\
    \midrule
    \multicolumn{1}{c}{\multirow{2}[2]{*}{Traffic light recognition}} & Zolfi et al. \cite{zolfi2021translucent} & Black & Disturb Camera Imaging & CVPR 2021 \\
    \multicolumn{1}{c}{} & B\&S \cite{patel2020bait} & White& Meaningful Patch & NeurIPS 2020\\
    \midrule
    \multicolumn{1}{c}{\multirow{5}[2]{*}{Vehicle detection}} & CAMOU \cite{zhang2019camou} & Black & Meaningless Patch & ICLR 2019 \\
    \multicolumn{1}{c}{} & ER \cite{wu2020physical} & Black & Meaningless Patch & ArXiv 2020 \\
    \multicolumn{1}{c}{} & UPC \cite{huang2020universal} & White & Meaningless Patch & CVPR 2020 \\
    \multicolumn{1}{c}{} & DAS \cite{wang2021dual} & White & Meaningful Patch & CVPR 2021 \\
    \multicolumn{1}{c}{} & DTA \cite{suryanto2022dta} & White & Meaningless Patch & CVPR 2022 \\
    \multicolumn{1}{c}{} & FCA \cite{wang2022fca} & White & Meaningless Patch & AAAI 2022 \\
    \multicolumn{1}{c}{} & ACTIVE \cite{suryanto2023active} & Black & Meaningless Patch & ICCV 2023\\
   
    \midrule
    \multicolumn{1}{c}{\multirow{2}[2]{*}{Road line detection}} & Boloor et al. \cite{boloor2020attacking} & Black & Meaningful Patch & JSA 2020 \\
    \multicolumn{1}{c}{} & Sato et al. \cite{sato2021dirty} & White & Meaningful Patch& US 2021 \\
    \midrule
   \multicolumn{1}{c}{\multirow{2}[2]{*}{Monocular Depth Estimation}}  & Yamanaka et al. \cite{yamanaka2020adversarial} & White & Meaningless Patch & IA 2020 \\
   \multicolumn{1}{c}{} & Cheng et al. \cite{cheng2022physical} & White & Meaningful Patch & ECCV 2022\\
    \multicolumn{1}{c}{} &  3D2fool \cite{zheng2024physical} & White & Meaningless Patch & CVPR 2024 \\
    \midrule
    \multicolumn{1}{c}{\multirow{5}[5]{*}{LiDAR perception}} & Sun et al. \cite{sun2020towards}  & Black & Adversarial Laser & US 2020 \\
    \multicolumn{1}{c}{} & Tu et al. \cite{tu2020physically} & White \& Black & Adversarial 3D-print & CVPR 2020 \\
    \multicolumn{1}{c}{} & Abdelfattah et al. \cite{abdelfattah2021towards} & White & Adversarial 3D-print & ICIP 2020 \\
     \multicolumn{1}{c}{} & FLAT \cite{li2021fooling} & White & Adversarial Spoofying & ICCV 2021 \\
      \multicolumn{1}{c}{} & AOA \cite{zhu2021can} & Black & Adversarial Laser & CCS 2021 \\
       \multicolumn{1}{c}{} & Sato et al. \cite{sato2022poster} & Black & Adversarial Laser & CCS 2022 \\
      \multicolumn{1}{c}{} & PRA \cite{cao2023you} & Black  & Adversarial Laser & US 2023 \\
     \multicolumn{1}{c}{} & PLA-LiDAR \cite{jin2023pla} & Black  & Adversarial Laser & SP 2023 \\
      \multicolumn{1}{c}{} & EVAA \cite{vishnu2023ev} & White & Adversarial Laser & TGRS 2023 \\
    \multicolumn{1}{c}{} & PCAP \cite{wang2023adversarial} & White & Adversarial Laser & TMM 2024 \\
    \multicolumn{1}{c}{} & A-HFR \cite{sato2025realism} & Gray & Adversarial Laser & NDSS 2025 \\

    \bottomrule
    \end{tabular}%
  \label{SD}%
\end{table*}%

% \begin{figure}[ht]
% \centering
% \includegraphics[width=0.6\textwidth]{advcam.png}       
% \caption{Compared  with AdvPatch \cite{brown2017adversarial} and RP2 \cite{evtimov2017robust}, AdvCam \cite{duan2020adversarial} is more natural for the human vision. This figure comes from \cite{duan2020adversarial}.}
% \label{advcam}
% \end{figure}

\subsubsection{\textbf{Self-driving}}
Ensuring people's safety is a prerequisite for the application of autonomous driving. A self-driving system needs to deal with complex situations in the real world. It must keep the car on the right road, avoid obstacles, pedestrians, and other vehicles along the way, and accurately identify traffic signs, lights, and lane lines. Physically adversarial attacks can pose threats in all aspects, challenging the fundamental safety principle. 

$\bullet$ \textbf{\emph{Traffic sign recognition.}}
Early physical attacks on traffic signs were mainly conducted in white-box settings. Lu et al. \cite{lu2017adversarial} introduced an attack on traffic sign detection, but it lacked robustness and required large perturbations. Since autonomous vehicles operate under varying distances, angles, and weather, robustness against real-world conditions is essential. The Expectation Over Transformation (EOT) technique addresses this by simulating diverse input conditions. Shapeshifter \cite{lu2017adversarial} improves attack success rates using iterative optimization and EOT. Unlike traditional digital EOT \cite{athalye2018synthesizing}, RP2 \cite{evtimov2017robust} directly captures physical variations, offering a more realistic approach. Eykholt et al. \cite{song2018physical} extended RP2 from classification to object detection.

Many methods apply transformations such as scaling, translation, rotation, and brightness changes, though their individual effects on robustness are often unclear. Sava et al. \cite{sava2022assessing} systematically evaluated 13 such transformations through physical experiments and proposed spatial heatmaps to visualize performance under varying distances and angles.

Despite improved robustness, many adversarial patterns remain visible. NatureAE \cite{xue2021naturalae} introduces adaptive masking to reduce perturbations while maintaining high success rates. AdvCam \cite{duan2020adversarial} uses neural style transfer to generate natural-looking corrosion patterns on stop signs. Sitawarin et al. \cite{sitawarin2018rogue} attack billboard images near traffic signs to mislead recognition systems. Zhong et al. \cite{zhong2022shadows} generate adversarial shadows via model queries, achieving high attack success with minimal perceptibility. IAP \cite{bai2021inconspicuous} progressively creates visually inconspicuous patches, evaluated through saliency maps and user studies.

Some works employ light to implement stealthy physical attacks. SLAP\cite{lovisotto2021slap} projects adversarial patches on traffic signs with light projectors, RFLA\cite{wang2023rfla} controls the reflected light color and shape, ILRs\cite{sato2024invisible} uses infrared light to disturb AI perception but are invisible to human eyes.

\begin{figure}[h]
    \centering
    \subfigure[RP2\cite{eykholt2018robust}]{
        \includegraphics[width=0.15\textwidth]{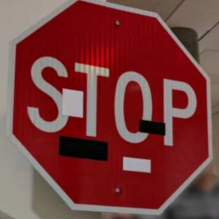}
    }
    \subfigure[AdvCAM\cite{duan2020adversarial}]{
        \includegraphics[width=0.15\textwidth]{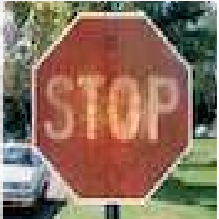}
    }
    \subfigure[SLAP\cite{lovisotto2021slap}]{
        \includegraphics[width=0.15\textwidth]{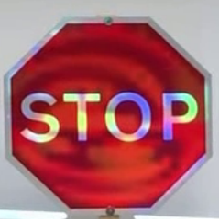}
    }
    \subfigure[RFLA\cite{wang2023rfla}]{
        \includegraphics[width=0.15\textwidth]{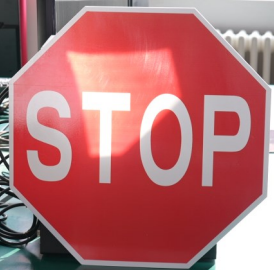}
    }
    \subfigure[ILRs\cite{sato2024invisible}]{
        \includegraphics[width=0.3\textwidth]{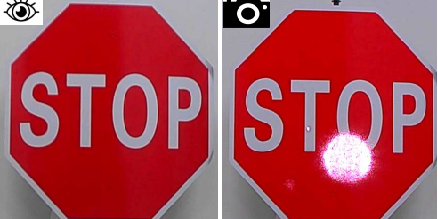}
    }
    \caption{Representative Methods of physically adversarial attacks in autonomous driving, traffic sign recognition. (a) Adversarial graffiti ; (b) Adversarial Camouflage; (c) Adversarial projection; (d) Reflected light attacks (e) Infrared light attacks.}
    \label{person_detection}
\end{figure}

$\bullet$ \textbf{\emph{Traffic light recognition.}}
Traffic light recognition systems for autonomous vehicles require fine-tuning when adapting to new cities. Patel et al. \cite{patel2020bait} proposed a physical poisoning attack during this fine-tuning process. They use a screen that displays a specific pattern based on the color of the traffic light, causing the DNN to learn incorrect correlations and make false predictions. Due to the difficulty of conducting such physical experiments, they implemented this method in a virtual environment instead. Recently, Zolfi et al. \cite{zolfi2021translucent} developed translucent stickers that create a local out-of-focus effect on the camera's optical lens, targeting Tesla's autopilot system. Compared to \cite{patel2020bait}, translucent patches are easier to implement and less likely to be detected.

$\bullet$ \textbf{\emph{Vehicle detection.}} Accurate and timely detection of surrounding vehicles is crucial for autonomous driving safety. Adversarial patterns can hide vehicles and threaten visual perception systems. Due to safety and cost constraints, most studies are conducted in virtual environments or with toy cars, evaluating robustness under various distances, angles, and lighting conditions \cite{zhang2019camou, wu2020physical, wang2021dual, suryanto2022dta, wang2022fca, suryanto2023active}. Zhang et al. \cite{zhang2019camou} use Unreal Engine 4 and gradient networks to create adversarial camouflage. Wu et al. \cite{wu2020physical} applied discrete search and enlarge-and-replicate methods to generate camouflage from small patches, both using black-box query attacks.

\begin{wrapfigure}{r}{0.65\textwidth}
  \centering
  \includegraphics[width=0.63\textwidth]{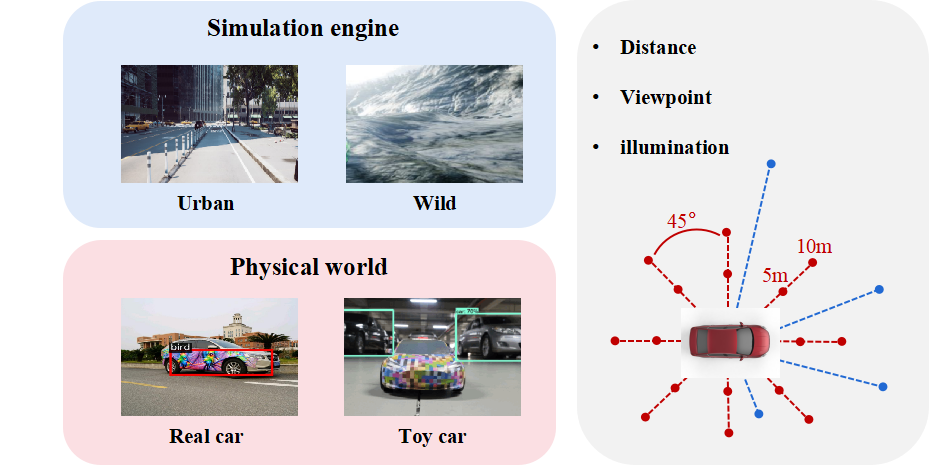}
  \caption{Vehicle detection attacks are conducted in simulators or real-world settings, requiring robustness across distance, viewpoint, and lighting.}
  \label{vehicle_detection/evaluation.png}
\end{wrapfigure}
In practice, query access is often unavailable, so surrogate models generate adversarial camouflage. Huang et al. \cite{huang2020universal} proposed Universal Perturbation Camouflage (UPC), attacking multiple detector components simultaneously. Wang et al. \cite{wang2021dual} developed Dual Attention Suppression (DAS), reducing model attention and diverting human focus, though its renderer has limitations. Wang et al. \cite{wang2022fca} introduced a differentiable rendering network for multi-view full-coverage camouflage (FCA). Suryanto et al. \cite{suryanto2022dta} used a differentiable transformation network (DTN) to simulate realistic vehicle features, but FCA showed limited real-world effectiveness, and DTN sometimes produced unnatural camouflage. To address this, Suryanto et al. \cite{suryanto2023active} proposed ACTIVE, which uses tri-planar mapping and regularization to improve robustness and stealthiness, achieving better performance across multiple tasks and physical scenarios.

$\bullet$ \textbf{\emph{Road line detection.}}  Automated Lane Centering (ALC) is a level-2 self-driving technique that stay centered in their lane. It is widely used in mainstream commercial models such as Tesla AutoPilot and SuperCruise. Early Tencent \cite{tencent2019experimental} paints fake road lines to fool Tesla Autopilot, but does not implement a drive-by test on the road with a lane line. Boloor et al. \cite{boloor2020attacking} proposed a query-based method that produces adversarial black lines to attack an autonomous simulator. Sato et al. \cite{sato2021dirty} proposed to produce a dirty road patch to deceive ALC systems. Their method is evaluated on OpenPliot and achieves 97.5\% ASR in 80 attack scenes. The mean success time is 1.6 seconds less than the average driver reaction time.

$\bullet$ \textbf{\emph{Monocular Depth Estimation.}}
Monocular Depth Estimation (MDE) plays a crucial role as an auxiliary component in fully vision-based autonomous driving systems (ADS), extending perception from 2D to 3D space. Yamanaka et al. \cite{yamanaka2020adversarial} initially targeted MDE using adversarial patches in white-box settings. Their objective was to optimize depth values within patch regions to specific depths, employing techniques from \cite{goodfellow2014explaining} to iteratively generate noticeable perturbations. In addition, they applied common transformation methods to enhance the robustness of these patches. Real-world experiments highlighted limited transferability between different MDE models \cite{lee2019big,guo2018learning}. To balance effectiveness and stealthiness, Cheng et al. \cite{cheng2022physical} proposed natural-style patches controlled by a differentiable mask for size adjustment. They assessed transferability across three MDE models and demonstrated impacts on downstream 3D reconstruction tasks. Zheng et al. \cite{zheng2024physical} introduced 3D2fool, an adversarial texture aimed at attacking MDE models. Their approach utilizes texture conversion (TC) techniques \cite{suryanto2022dta} to generate object-agnostic textures, enabling applicability to various types of vehicles. These 2D textures are then converted into 3D views, with the Differential Transformation Network (DTN) \cite{suryanto2022dta} used to render mimic patterns. Compared to traditional 2D patches \cite{yamanaka2020adversarial,cheng2022physical,guesmi2024saam,guesmi2023aparate}, 3D2fool demonstrates enhanced robustness against adverse weather conditions.

\begin{figure}[h]
    \centering
    \subfigure[Tu et al. \cite{tu2020physically}]{
        \includegraphics[width=0.3\textwidth]{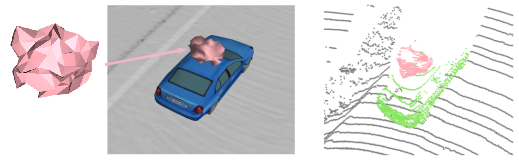}
    }
    \subfigure[AOA \cite{zhu2021can}]{
        \includegraphics[width=0.3\textwidth]{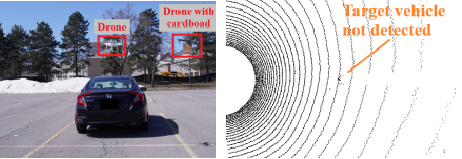}
    }
    \subfigure[FLAT \cite{li2021fooling}]{
        \includegraphics[width=0.3\textwidth]{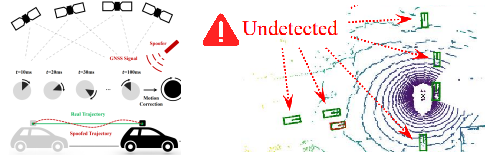}
    }
    \caption{Representative physical attacks in LiDAR perception-based autonomous driving. (a) Adversarial meshes; (b) Reflected laser by controlling drones; (c) Spoof GNSS signal to disturb cars' trajectory.}
    \label{lidar_perception}
\end{figure}

Besides visual autopilot, some modern autonomous driving systems use LiDAR perception, which is harder to attack with image-level adversarial examples. Two main physical attack methods target LiDAR: creating 3D adversarial meshes \cite{tu2020physically, abdelfattah2021towards} and using adversarial lasers to disrupt scanning \cite{sun2020towards, li2021fooling}. Tu et al. \cite{tu2020physically} generated adversarial meshes in both white-box and black-box settings by fitting the mesh to vehicle roofs and adjusting vertices to lower detection confidence. Abdelfattah et al. \cite{abdelfattah2021towards} created multimodal attacks by rendering 3D meshes into both point clouds and 2D images. However, manufacturing precise 3D meshes is challenging, and some adversarial features may be lost when sampled as point clouds. Zhu et al. \cite{zhu2021can} proposed a flexible physical attack using drones and reflective surfaces to interfere with LiDAR lasers, which is easier to implement and covert.

Other studies \cite{cao2019adversarial,petit2015remote,shin2017illusion,sun2020towards,li2021fooling} show LiDAR systems are vulnerable to spoofing lasers that create fake obstacles or hide real ones. Sun et al. \cite{sun2020towards} present the first black-box spoofing attack manipulating lasers to generate occlusions far from the vehicle. Wang et al. \cite{wang2023adversarial} improved this with polar coordinate perturbations for higher attack success with fewer points. Li et al. \cite{li2021fooling} proposed a trajectory spoofing attack (FLAT) that deceives motion correction to cause vehicle deviation. Cao et al. \cite{cao2023you} exploited LiDAR sensor transformations and filtering to launch effective attacks on detectors and fusion models. Sato et al. \cite{sato2022poster} built optical and electronic interference systems for real-world evaluations, and Jin et al. \cite{jin2023pla} developed a LiDAR transmitter injecting many more spoof points, enhancing attack power.

In summary, adversarial lasers are more practical for LiDAR attacks, while 3D meshes allow multimodal attacks affecting both LiDAR and visual systems.

\begin{table*}[htbp]
  \centering
  \caption{Representative physical attacks against \textbf{security surveillance systems}}
    \begin{tabular}{ccccc}
    \midrule
    \multicolumn{1}{c}{Sub-tasks} & Methods & \multicolumn{1}{c}{Settings} & Physical Forms & \multicolumn{1}{c}{Sources} \\
    \midrule
    \multirow{7}[2]{*}{Person detection} 
    & Thys et al. \cite{thys2019fooling} & \multicolumn{1}{c}{White} & Meaningless Patch (Cardboard) & CVPR-W 2019  \\
    \multicolumn{1}{c}{} & UPC \cite{huang2020universal} & \multicolumn{1}{c}{White} & Meaningless Patch (Clothes) & CVPR 2020 \\
    \multicolumn{1}{c}{} & Adv-Cloak \cite{wu2020making} & \multicolumn{1}{c}{Black} & Meaningless Patch (Clothes) & ECCV 2020 \\
    \multicolumn{1}{c}{} & Adv T-shirt \cite{xu2020adversarial} & \multicolumn{1}{c}{White} & Meaningless Patch (Clothes) & ECCV 2020 \\
    \multicolumn{1}{c}{} & Naturalistic Patch \cite{hu2021naturalistic} & \multicolumn{1}{c}{White} & Meaningful Patch (Clothes) & ICCV 2021 \\
    \multicolumn{1}{c}{} &  Adv-Bulbs \cite{zhu2021fooling} & \multicolumn{1}{c}{White} & Adversarial Light (Infrared) & AAAI 2021 \\
    \multicolumn{1}{c}{} &  IIC \cite{zhu2022infrared} & \multicolumn{1}{c}{White} & Meaningless Patch (Infrared) & CVPR 2022 \\
    \multicolumn{1}{c}{} &  AIP \cite{wei2023physically} & \multicolumn{1}{c}{White} & Meaningless Patch (Infrared)  & CVPR 2023 \\
    \multicolumn{1}{c}{} &  AdvCaT \cite{hu2023physically} & \multicolumn{1}{c}{White} & Meaningful Patch (Clothes)  & CVPR 2023 \\
    
    \multicolumn{1}{c}{} &  DAP \cite{guesmi2024dap} & \multicolumn{1}{c}{White} & Meaningful Patch (Clothes)  & CVPR 2024 \\
    \multicolumn{1}{c}{} &  Wei et al. \cite{wei2023unified} & \multicolumn{1}{c}{Black} & Meaningless Patch (Cross-modal)  & ICCV 2024 \\
    \midrule
    Person re-ID & Wang et al. \cite{wang2019advpattern}  & \multicolumn{1}{c}{White} & Meaningless Patch (Clothes) & ICCV 2019 \\
    \midrule
    \multirow{2}[2]{*}{Person tracking} & PAT \cite{wiyatno2019physical} & \multicolumn{1}{c}{White} & Meaningless Patch (Poster) & ICCV 2019 \\
    \multicolumn{1}{c}{} & Ding et al. \cite{ding2021towards} & \multicolumn{1}{c}{White} & Meaningless Patch (Clothes) & AAAI 2021 \\
    \midrule
    Action recognition & Pony et al. \cite{pony2021over} & White & Adversarial Light & CVPR 2021 \\
    \midrule
    \end{tabular}%
  \label{SS}%
\end{table*}%

\subsubsection{\textbf{Security surveillance}}
To enhance public safety, intelligent security monitoring is extensively employed in scenarios such as crime tracking and identifying illegal behavior. However, criminals can use physically adversarial examples to evade security surveillance, posing potential security threats.

$\bullet$ \textbf{\emph{Person detection.}} Person detection is essential for security surveillance and relies on visible-light and infrared imaging to handle varying lighting conditions. Early visible-light attacks use printed adversarial patterns to evade detection \cite{thys2019fooling}, and virtual datasets were developed to control testing conditions \cite{huang2020universal}. Transferability of attacks across detection systems has been widely studied \cite{wu2020making,xu2020adversarial,hu2021naturalistic,hu2022adversarial,zhu2021fooling,zhu2022infrared}. Robustness to clothing deformation and viewpoint changes is improved by mapping adversarial patterns onto 3D clothing via Thin Plate Spline and topological projection \cite{wu2020making, xu2020adversarial, hu2022adversarial, hu2023physically}. Naturalistic patterns are generated using BigGAN \cite{hu2021naturalistic}, while dynamic and inconspicuous patches are created using Creases Transformation to reduce perceptibility \cite{guesmi2024dap}.

In infrared detection, bulb arrays optimize light distribution to interfere with sensors efficiently \cite{zhu2021fooling}. Infrared invisibility clothing made from thermal insulation materials produces high-contrast patterns under thermal cameras \cite{zhu2022infrared}, though implementation is challenging. Optimizable infrared adversarial patches enhance manufacturability and robustness to angle and distance variations \cite{wei2023physically}.
Single-modal patches have a limited effect due to multi-sensor systems. Cross-modal adversarial patches optimized by Differential Evolution are designed to remain covert in infrared while fooling visible-light detectors \cite{wei2023unified}.

\begin{figure}[h]
    \centering
    \subfigure[Adv T-shirt \cite{xu2020adversarial}]{
        \includegraphics[width=0.15\textwidth]{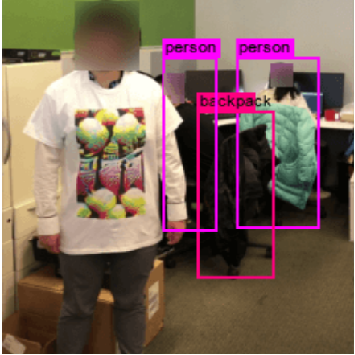}
    }
    \subfigure[DAP \cite{guesmi2024dap}]{
        \includegraphics[width=0.15\textwidth]{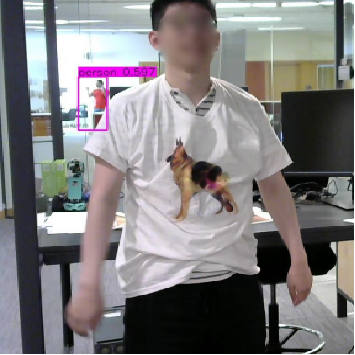}
    }
    \subfigure[AdvCaT \cite{hu2023physically}]{
        \includegraphics[width=0.15\textwidth]{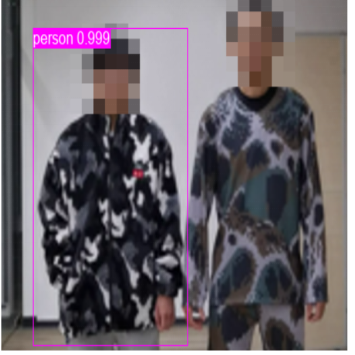}
    }
    \subfigure[AIP \cite{wei2023physically}]{
        \includegraphics[width=0.15\textwidth]{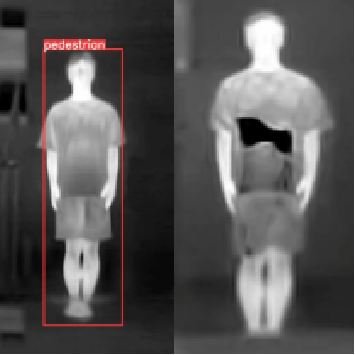}
    }
    \subfigure[IIC \cite{zhu2022infrared}]{
        \includegraphics[width=0.15\textwidth]{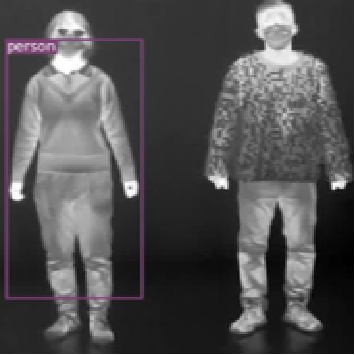}
    }
    \subfigure[Wei et al. \cite{wei2023unified}]{
        \includegraphics[width=0.15\textwidth]{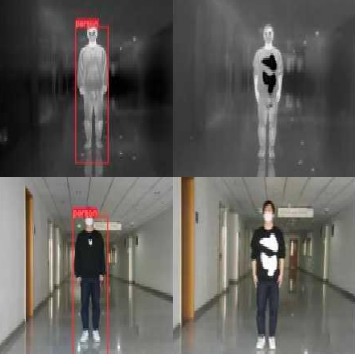}
    }
    \caption{Representative Methods of physically adversarial attacks against person detection. (a) Adversarial T-shirt; (b) Dynamic adversarial patch; (c) Adversarial camouflage textures; (d) Infrared invisible patches; (e) Infrared invisible clothes; (f) Cross-modal invisible patch (Visible light and Infrared light).}
    \label{person_detection_fig}
\end{figure}

$\bullet$ \textbf{\emph{Person re-ID.}}
Person re-identification (re-ID) is a critical task in security surveillance, serving as a sub-problem of image retrieval aimed at matching a person of interest across different camera views. Due to variations in shooting angles and distances, changes in attire, and differing environmental conditions, attacking re-ID models presents a significant challenge. Wang et al. \cite{wang2019advpattern} introduced the first physical attack against re-ID models. They implemented dodging attacks to cause identification mismatches and impersonation attacks to mimic a target person. In their physical experiments, they created adversarial clothing and tested various shooting angles and distances. The results demonstrated a 60\% drop in matching precision under dodging attacks and a 67.9\% mean Average Precision (mAP) in impersonation attacks.

$\bullet$ \textbf{\emph{Person tracking.}}
Unlike person detection, which involves identifying individuals, person tracking requires quick responses to a moving person and is usually modeled as a similarity matching problem \cite{bertinetto2016fully, li2018high}. Previous attacks on single object tracking have mostly been implemented in the digital domain \cite{chen2020one, yan2020hijacking, wu2019sta}, and do not transfer well to physical settings. Wiyatno and Xu \cite{wiyatno2019physical} proposed a method of generating adversarial textures to be displayed on a screen or poster. However, this approach requires a large area of perturbations covering the background, making it impractical for outdoor tracker attacks. Leveraging the spatial texture stationarity in raw images \cite{gatys2016image, kurakin2016adversarial}, Ding et al. \cite{ding2021towards} introduced Maximum Textural Discrepancy (MTD) to maximize the discrepancy between the template image and the search image. In physical attacks, they conducted experiments on state-of-the-art trackers (SiamRPN++ \cite{li2019siamrpn++}, SiamMask \cite{wang2019fast}) and achieved over 40\% ASR.

$\bullet$ \textbf{\emph{Action Recognition.}}
The action recognition model takes a time-series image as input. Thus the attacker needs to update adversarial examples in each frame, but physically adversarial examples are difficult to change immediately, such as adversarial patches. Pony et al. \cite{pony2021over} proposed an implementable approach to misleading action recognition by controlling the LED light. Instead of computing the perturbation for each pixel, they add an RGB offset with the same value for each frame.

\subsubsection{\textbf{Remote sensing}}
Remote sensing is closely related to public safety and national security. Remote sensing images are captured by drones or satellites, with image quality often challenged by atmospheric conditions and the distance between the camera and objects. Adversarial patches offer a practical solution to address these challenges. Czaja et al. \cite{czaja2018adversarial} conducted attacks on aerial image classifiers and discussed the potential impact of atmospheric conditions. Hollander et al. \cite{den2020adversarial} attempted to camouflage large military assets such as aircraft and warships, though they did not implement physical patches in real-world scenarios. Du et al. \cite{du2022physical} were the first to implement physical attacks on aerial images, applying patches not only on vehicles themselves but also in the vicinity of vehicles. Moreover, they considered changes in weather and seasons, simulating different conditions to enhance the robustness of adversarial patches.

\subsubsection{\textbf{Embodied agent}}
 With the advancement of multi-modal techniques and virtual environments, embodied agents are employed to simulate real interactions with humans and their surroundings. Embodied Question Answering (EQA) \cite{das2018neural} and Embodied Vision Recognition (EVR) \cite{yang2019embodied} are utilized to enable speech and visual interactions. Liu et al. \cite{liu2020spatiotemporal} proposed a novel spatiotemporal attack on embodied agents. In this attack scenario, the agent locates a chessboard in various rooms to answer a question, while the attacker generates perturbations on the target 3D object to deceive the agent (these local perturbations can manifest as patches in the physical world). Results demonstrate that their method significantly reduces accuracy in question answering (from 40\% to 5\% on average) and visual recognition (from 89.91\% to 18.32\%).

\subsection{\textbf{Attack forms}}
Before implementing a physical attack, the adversarial example needs to be appropriately manufactured. Attackers often focus on the feasibility of a method in real environments, considering factors such as resistance to adverse environmental effects, ease of manufacture, and preventing the adversarial pattern from being detected by the human eye. This section introduces various forms of attacks, including adversarial patches, optical adversarial attacks, and 3D-printed adversarial objects.

\subsubsection{\textbf{Adversarial patches}}
The adversarial patch is the most common approach in physical attacks. While digital attacks generate perturbations across the entire image, which is impractical for real-world implementation, adversarial patches only modify local pixels. These patches can be easily printed and directly applied to the target. A mask is typically use to control the shape of the perturbed area. Once the adversarial patch is optimized in the digital domain, it is crafted and placed on the surface of the object. Adversarial patches can be divided into meaningless patches and meaningful patches according to the context of the adversarial pattern.

% The adversarial patch can be defined as:
% \begin{equation}
% I_{adv}=M\odot I(x_{patch})+(\textbf{1}-M)\odot I,
% \end{equation}
% where I is the clean object, and $I(x_{patch})$ denotes the adversarial patch. $M$ is the mask, $M_{i,j}\in\{0,1\}^d$, where \textbf{1} is a unit matrix which has the same size as M.

% \begin{figure}[htbp]
% \centering
% \subfigure[Adv-Glass \cite{sharif2016accessorize} vs. Adv-Sticker \cite{wei2022adversarial}.]{
% \includegraphics[width=5cm]{patch1.png}
% }
% \quad
% \subfigure[AdvTexture \cite{hu2022adversarial} vs. Naturalistic Patch \cite{hu2021naturalistic}.]{
% \includegraphics[width=5cm]{patch2.png}
% }
% \quad
% \subfigure[RP2 \cite{evtimov2017robust} vs. Adv-Shadow \cite{zhong2022shadows}.]{
% \includegraphics[width=5cm]{patch3.png}
% }
% \quad
% \subfigure[DTA \cite{suryanto2022dta} vs. DAS \cite{wang2021dual}.]{
% \includegraphics[width=5cm]{patch4.png}
% }
% \caption{Visual difference between meaningless patches (Left) and meaningful patches (Right).}
% \label{patch}
% \end{figure}

\textbf{\emph{(1) Meaningless patches.}}
From a human observation standpoint, meaningless patches do not correspond to real-world objects. Generating such patches requires calculating perturbed values for each pixel. In the early stages, Karmon et al. \cite{karmon2018lavan} proposed Localized and Visible Adversarial Noise (LaVAN) to facilitate physical implementations. Brown et al. \cite{brown2017adversarial} proposed creating adversarial patches for real-world scenarios. To make adversarial patches look smoother, attackers minimize the TV loss \cite{strong2003edge} to reduce the variations between adjacent pixels. Digital adversarial examples can display rich colors. For a real-world attack, printer colors are limited. Sharif et al. \cite{sharif2016accessorize} first proposed a non-printed score (NPS) to eliminate printed color error, which makes the physically adversarial patches closer to the digital patches. 

% NPS is formulated as:
% \begin{equation}
%  \mathcal{L}_{nps}=\sum\limits_{\hat{p} \in \hat{P}}\prod \limits_{p \in P}|\hat{p}-p|,
% \end{equation}

% where $P\subset{[0,1]^3}$ is the set of printable RGB space, $\hat{P}$ is the set of the pixel value in adversarial pattern, $\hat{p}$ represents one pixel's RGB value of the adversarial pattern, p is the printable color and $p\in P$. 

Adversarial patches are pasted on the surface of 3D objects, such as human faces, cars, and clothes. These curved surfaces might destroy the effectiveness of the adversarial pattern, so attackers must model deformations. For the human face, Pautov et al. \cite{pautov2019adversarial} train a grid generator to simulate Non-linear transformation, then project the adversarial pattern on the grid. Komkov et al. \cite{komkov2021advhat} used STL (spatial transformation layer) to project the perturbation on a hat. STL includes two steps: off-plane bending and pitch rotation. To realize wearable adversarial examples, Wu et al. \cite{wu2020making} and Xu et al. \cite{xu2020adversarial}  modeled non-rigid deformations by TPS \cite{inproceedings}. \cite{xu2020adversarial} showed that the deformation modeling can significantly improve the ASR of 48\%-74\% against YOLOv2 and 34\%-61\% against Faster-RCNN \cite{ren2015faster}. Hu et al. \cite{hu2022adversarial} introduced the Toroidal Cropping (TC) technique, which can project points from a two-dimensional plane onto a torus by two folds. The best texture is searched by random cropping at the junction of the recurring patch plane.

\textbf{\emph{(2) Meaningful patches.}}
A patch that humans can recognize as a real object is meaningful. Most existing meaningful patches \cite{sharif2019general,hu2021naturalistic,xiao2021improving,yin2021adv} are generated by GAN. Adversaries generally train a generator to produce realistic or natural-style adversarial patches \cite{xue2021naturalae,duan2020adversarial,cheng2022physical}. Face stickers are common every day and can be seen at festivals or large events. Recently, Wei et al. \cite{wei2022adversarial} optimized cartoon stickers' location and rotation angle to achieve a high success rate against the face recognition system. They do not need to compute each pixel value of adversarial patterns. Thus their method is easier to implement and more threatening. Inspired by natural shadows, Zhong et al. \cite{zhong2022shadows} produced adversarial shadows in the real world and achieve a 95\% success rate. They construct a triangular area and use the PSO strategy to search for the optimal location by optimizing the three coordinates of the vertices. In order to adjust the brightness of shadows, they manipulate the L channel of LAB space according to the SBU Shadow dataset \cite{vicente2016large}, and they statistically analyze the mean ratio of pixel values of LAB triple channels to simulate settings of realistic shadows. However, DAS \cite{wang2021dual} combines the meaningful patch and the meaningless patch. The patch generated by DAS consists of two parts: the outline of a smiling face and the texture of illegible patterns. Neural networks prefer texture information when extracting features, so the adversarial texture is used to deceive the attention of the DNN. At the same time, a meaningful contour is used to distract human attention, which makes the adversarial texture inconspicuous.

% Usually, 3D objects with an only outer surface can be successfully attacked. Objects with an inner surface (such as cups) are difficult to perceive, because adversarial point clouds may aggregate on the inner surface.

\subsubsection{\textbf{Optical adversarial attacks}}
CV systems require optical perception modules, creating a vulnerability to physically adversarial attacks. Adversaries can exploit the imaging principles of cameras and the characteristics of image sensor processors to launch attacks. Based on existing research, we categorize Optical Adversarial Attacks into two types: adversarial lights and disturb camera imaging.

% \begin{figure*}[ht]
% \centering
% \includegraphics[width=0.8\textwidth]{projection.png}        
% \caption{The framework of adversarial light projection. Due to the background lighting, The table shows only projecting the adversarial illumination with environment loss compensation can successfully fool the VGG-16. This figure is from \cite{gnanasambandam2021optical}.}
% \label{projection}
% \end{figure*}

\begin{figure}[h]
    \centering
    \subfigure[Adversarial Projection \cite{gnanasambandam2021optical}.]{
        \includegraphics[width=0.3\textwidth]{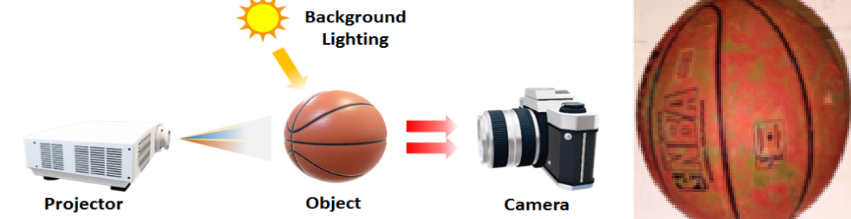}
    }
    \subfigure[Adversarial Point Laser \cite{duan2021adversarial}.]{
        \includegraphics[width=0.3\textwidth]{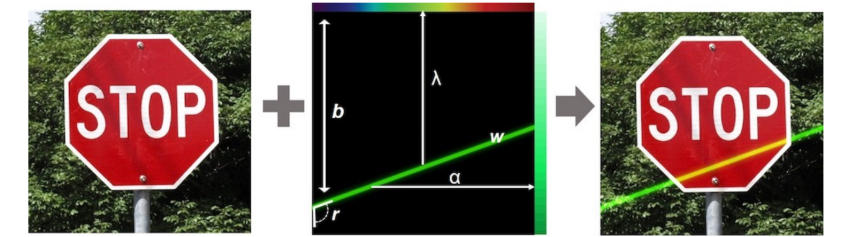}
    }
    \subfigure[RSE Adversarial light \cite{sayles2021invisible}.]{
        \includegraphics[width=0.3\textwidth]{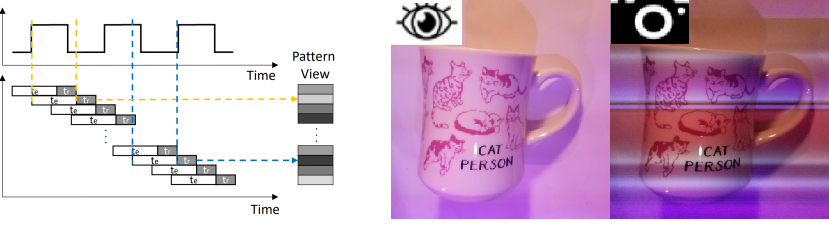}
    }

    \caption{Representative Methods in adversarial lights. }
    \label{adversarial_light}
\end{figure}

 \textbf{\emph{(1) Adversarial lights.}}
A simple way to extend adversarial patches from printed images to visible light is by projecting light. Nguyen et al. \cite{nguyen2020adversarial} proposed a real-time physical attack by projecting adversarial patterns onto faces after calibrating camera and projector settings, including position and color calibration. They also identify factors that can cause failures, such as strong ambient light and poor face poses. Lovisotto et al. \cite{lovisotto2021slap} created Short-Lived Adversarial Perturbations (SLAP) by projecting RGB light onto stop signs, using frame interpolation to handle lighting changes and reduce camera noise, achieving high attack success against defenses like SentiNet \cite{chou2020sentinet}. Gnanasambandam et al. \cite{gnanasambandam2021optical} modeled image transformations from the projector to the camera and considered background illumination to keep adversarial lights effective in different environments. Some attacks avoid special equipment like projectors, making outdoor use easier. Duan et al. \cite{duan2021adversarial} used laser pointers with optimized parameters to launch physical attacks. Wang et al. \cite{wang2023rfla} proposed the Reflected Light Attack (RFLA), controlling reflected light spots with mirrors, filters, and shaped cutouts to fool traffic sign recognition. For stealth, infrared light sources are invisible to humans but can be detected by cameras. Sato et al. \cite{sato2024invisible} used infrared laser reflections optimized for size, intensity, and position to attack traffic sign systems. Sayles et al. \cite{sayles2021invisible} exploited the rolling shutter effect in cameras to create invisible stripe patterns that mislead classifiers, using modulated RGB LED illumination beyond human perception.

 \textbf{\emph{(2) Disturb camera imaging.}}
The camera lens and image sensor processor (ISP) play important roles in the final image quality, so they can be used as a backdoor to implement the physical attack. Disturb camera imaging specifically exploits the principles of optical imaging rather than disrupting the functions and structure of hardware.

\begin{figure}[h]
    \centering
    \subfigure[Translucent dots.]{
        \includegraphics[width=0.2\textwidth]{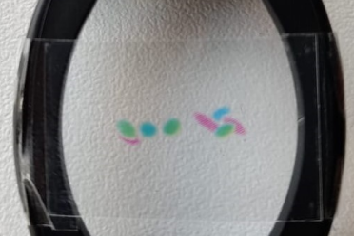}
    }
    \subfigure[Translucent dots on lense.]{
        \includegraphics[width=0.2\textwidth]{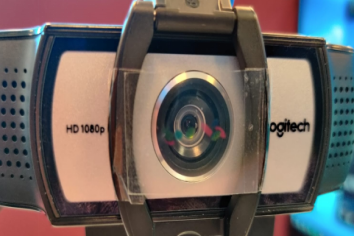}
    }
    \subfigure[Disturbed image.]{
        \includegraphics[width=0.205\textwidth]{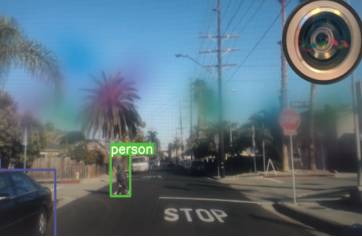}
        \label{adversarial_light(c)}
    }
    \subfigure[Disturbed Tesla ADAS.]{
        \includegraphics[width=0.2\textwidth]{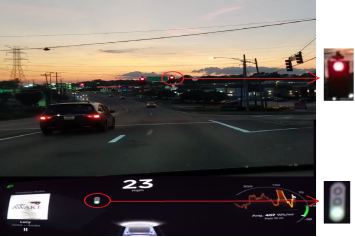}
        \label{adversarial_light(d)}
    }

    \caption{Adversarial translucent sticker on the camera lense\cite{zolfi2021translucent}. The functions of traffic sign recognition (c) and light recognition (d) are disturbed.}
    \label{adversarial_light_figs}
\end{figure}

Based on image formation principles, an Image Signal Processor (ISP) converts raw images to RGB. Phan et al. \cite{phan2021adversarial} designed a neural approximation of the ISP and generated perturbations via multi-task optimization to deceive specific camera ISPs. Their method, leveraging shared ISP components like demosaicing, generalizes well to other black-box imaging systems.

Li et al. \cite{li2019adversarial} proposed using a translucent sticker (TS) on the camera lens to introduce adversarial blur patterns. They modeled this effect digitally, used SSIM to guide optimization, and applied color mapping with greedy search and gradient descent. A 10-dot TS achieved nearly 50\% ASR against ResNet50. Zolfi et al. \cite{zolfi2021translucent} extended this to object detectors, fooling Tesla’s ADAS into misclassifying red lights as green in real-world driving.

\begin{figure}[ht]
\centering
\includegraphics[width=0.9\textwidth]{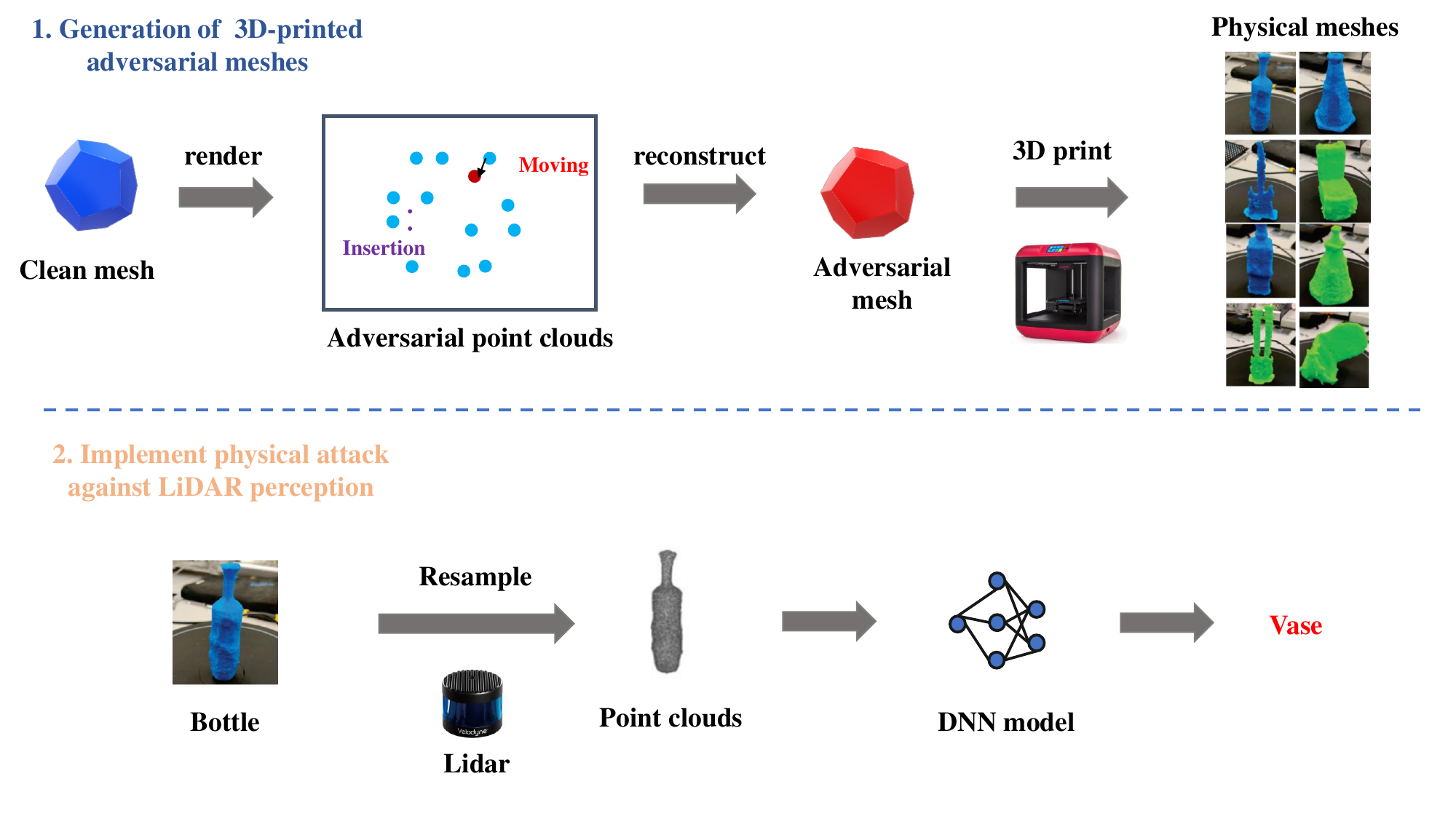} 
\vspace{-0.6cm}
\caption{3D-printed adversarial attack including two steps: (1) Generation of 3D-printed adversarial meshes. Firstly, a clean mesh is rendered to point clouds. Secondly, add an offset to a point or insert new points and optimize parameters to achieve an attack in the digital domain. Thirdly, reconstruct point clouds into adversarial meshes. Finally, print 3D objects according to adversarial meshes; (2) Implement a physical attack against LiDAR perception. 3D-printed objects will be resampled by LiDAR sensors, which may drop some adversarial point clouds. Pictures of 3D-print objects are from \cite{tsai2020robust}.}
\label{3D print}
\end{figure}

\subsubsection{\textbf{Adversarial 3D-print}}
Although physical attacks on 2D images have exposed severe security problems, little research has been reported on 3D perception. For the 3D adversarial attack, the coordinate changes to a triplet, which increases optimization's complexity and challenges crafting physical forms. A 
framework of adversarial 3D-print is shown in Fig.~\ref{3D print}. Tsai et al. \cite{tsai2020robust} put forward a novel approach to creating printable adversarial objects in the physical world. They print 3D objects after producing adversarial examples in the digital domain, but attacks mostly fail for complex objects during re-sampling, which can drop the information of perturbed point clouds. Wen et al. \cite{wen2020geometry} proposed a Geometry-Aware Adversarial Attack ($GeoA^3$) and crafted 3D-printed adversarial objects. To better preserve adversarial effects when re-sampling point clouds from the surface meshes, $GeoA^{3}$-IterNormPro is used to optimize perturbations via iterative normal projection. Compared to \cite{tsai2020robust} in physical attacks, \cite{wen2020geometry} shows a better result in different models. 

In addition to LiDAR perception, some works focus on fooling image recognition models by changing texture or geometry to achieve adversarial attacks \cite{athalye2018synthesizing, zeng2019adversarial, oslund2022multiview, xiao2019meshadv}. Despite their greater threat, research on targeted 3D adversarial attacks in the physical world is still lacking. Huang et al. \cite{huang2024towards} addressed this gap by proposing a physically untargeted transferable 3D adversarial attack (TT3D). Unlike existing methods that directly alter vertex colors \cite{suryanto2022dta, xiao2019meshadv}, TT3D creates adversarial texture objects using a grid-based Neural Radiance Fields (NERF) approach. This involves dual optimization to simultaneously update the parameters of the appearance feature grid and the MLP. Results show that TT3D exhibits prior transferability across multiple models and tasks (classification, detection, and image captioning) and demonstrates effectiveness in the physical world.

\subsection{\textbf{Attack methods}}
\textcolor{black}{Based on the level of knowledge about the target model or training dataset, physical attacks can be categorized into three settings: white-box, black-box, and gray-box. In a white-box setting, the attacker has full access to the model architecture, parameters, and training data. This allows for the most optimized and effective attack generation, as gradients and internal representations can be directly exploited. In a black-box setting, the attacker has no internal knowledge of the model and can only observe the input-output behavior. Attacks in this setting often rely on query-based strategies or surrogate models to approximate the target system. The gray-box setting lies between these two extremes. The attacker has partial knowledge—for example, access to the model architecture but not the parameters, or access to a similar dataset but not the exact training data.} 

\subsubsection{\textbf{White-box attacks}}
We introduce the white-box attacks in the physical attacks from gradient-based methods and optimization-based methods, respectively.

\textbf{\emph{(1) Gradient-based attacks.}}
Most white-box attacks generate perturbations based on the gradient information of the target model\cite{wiyatno2019physical,madry2018towards,dong2018boosting,xie2019improving}. In most physical cases, parameters can be optimized directly with gradient descent in the white-box setting. Singh et al. \cite{singh2021brightness} proposed to generate robust adversarial examples against light changes by exploiting nonlinear brightness transformation and PGD attack. However, sometimes attackers must establish differentiable models themselves so that the gradient propagates in the framework of producing adversarial examples \cite{gnanasambandam2021optical,suryanto2022dta,phan2021adversarial,sayles2021invisible}. The gradient-based method can achieve a high success rate in white-box settings, but their performances usually drop when transferred to other black-box models. 

\textbf{\emph{(2) Optimization-based attacks.}}
Gradient-based attacks often fail against models with defenses like adversarial training \cite{madry2018towards} and knowledge distillation \cite{papernot2016distillation}, making optimization-based attacks a better choice. Carlini \& Wagner (C\&W) \cite{carlini2017towards} framed adversarial example generation as a box-constrained optimization problem and showed it could bypass KD and other defenses. Their method inspired many follow-up attacks. Eykholt et al. \cite{evtimov2017robust} added EOT and NPS to C\&W to attack traffic sign classifiers. Yang et al. \cite{yang2020beyond} applied C\&W to SSD with color and brightness constraints, using logits distance as the objective. Pony et al. \cite{pony2021over} revised the C\&W loss with gradient-smoothing terms and temporal regularization for imperceptibility. Tsai et al. \cite{tsai2020robust} extended the approach to point clouds and physical objects, using Chamfer distance to measure perturbations. While C\&W is effective across visual tasks, its adversarial examples still lack transferability and stealth.

\subsubsection{\textbf{Black-box attacks}}
In the physical world, adversaries can only access limited information about target models. Therefore, it is more valuable to study attack methods under black-box settings. In this part, we divide black-box attacks into query-based, hybrid-based, and transfer-based attacks.

\textbf{\emph{(1) Query-based attacks.}}
For the query-based attack, we assume that the training data and the target model are unknown but allow the attacker to obtain the output of the target model, such as probability or class. To address the challenges of black-box and physical attacks, several optimization-based strategies have been proposed. Wei et al. \cite{wei2022adversarial} designed a physical attack by adjusting the sticker’s position and rotation using an evolutionary algorithm, specifically a region-based heuristic differential evolution (RHDE) method, which efficiently searches near promising solutions and is robust to inaccurate placement. Reinforcement learning (RL) has also been applied: Guo et al. \cite{wei2021generating} developed an RL framework to jointly optimize patch location and perturbation, significantly reducing query counts compared to methods like RHDE and ZO-AdaMM \cite{chen2019zo}, while maintaining high attack success on popular face recognition models. Sharif et al. \cite{sharif2016accessorize} utilized particle swarm optimization (PSO) for impersonation attacks against Face++, introducing a recursive strategy to overcome limitations in top-k prediction outputs. Zhong et al. \cite{zhong2022shadows} similarly adopted PSO after encountering instability with gradient-based methods when generating adversarial shadows for traffic sign recognition. Duan et al. \cite{duan2021adversarial} proposed a greedy approach to tune physical parameters of adversarial laser beams (AdvLB), such as wavelength and intensity, within feasible ranges. Furthermore, Wei et al. \cite{wei2023distributional} introduced a distribution-based optimization method, treating effective attack locations as a learnable distribution rather than individual points, which improves transferability and reduces queries in attacking unseen face recognition models.

 \textbf{\emph{(2) Transfer-based attacks.}}
For transfer-based attacks, we assume that the target model is unknown, but all or part of the training data can be obtained. Adversaries train a surrogate model to generate adversarial examples and implement attacks against the target model based on the transferability of the adversarial example. To improve the transferability and physical effectiveness of adversarial examples, various advanced methods have been proposed. Ensemble attacks leverage multiple models during optimization to enhance generalization. For instance, Wu et al. \cite{wu2020making}, Xu et al. \cite{xu2020adversarial}, Hu et al. \cite{hu2021naturalistic}, and Zhu et al. \cite{zhu2021fooling} designed ensemble loss functions—such as summing detector losses or using min-max formulations with domain weights—to balance attack strength across different models. Meanwhile, generative approaches use GANs to create diverse and realistic adversarial examples with limited data. Sharif et al. \cite{sharif2019general} generated adversarial patches using GANs to perform impersonation attacks on both digital and physical face recognition systems. Kong et al. \cite{kong2020physgan} developed PhysGAN for generating robust roadside posters that mislead autonomous driving models. Jan et al. \cite{jan2019connecting} simulated the digital-to-physical (D2P) process with conditional GANs and EOT, while Bai et al. \cite{bai2021inconspicuous} used attention maps and coarse-to-fine perturbation to target vulnerable regions, achieving high transferability on real-world traffic signs. To further boost generalization, meta-learning is introduced. Feng et al. \cite{feng2021meta} proposed a class- and model-agnostic meta-learning (CMML) approach to train robust generators under D2P transformations. Yin et al. \cite{yin2021adv} used meta-learning to create transferable adversarial makeup, optimizing across multiple face recognition models and achieving superior attack success rates on commercial platforms like Face++ and Microsoft.

\textcolor{black}{\textbf{\emph{(3) Hybrid-based attacks.}}
In hybrid attack strategies, adversaries first build a surrogate white-box model by stealing the target model through black-box queries, then generate physical adversarial examples based on this surrogate. Although end-to-end physical demonstrations are rare, studies on model stealing and transferability show this approach is both feasible and impactful. The first stage—model stealing—aims to replicate the target model with limited queries. Hondru et al. \cite{hondru2023towards} proposed a few-shot method combining diffusion-generated proxy images, active sample selection, and self-paced pseudo-labeling to improve surrogate accuracy under tight query budgets. Zhu et al. \cite{zhu2025efficient} introduced E3, a data-free extraction method using public OOD data and language-model-guided queries, enhanced with adaptive training techniques like TTDA, achieving high performance with minimal queries. Liu et al. \cite{liu2023shrewdattack} presented ShrewdAttack, which leverages pre-trained models and confidence-based query partitioning to efficiently extract MLaaS models, reaching 96\% accuracy on Azure with only 5–7\% of the original data, further confirming the real-world viability of surrogate-based attacks.}

\textcolor{black}{After successfully stealing a model, the attacker obtains a white-box surrogate that can be used to create adversarial examples. This approach relies on the transferability principle, where attacks crafted on one model also work on others. Research on robust physical adversarial attacks using known or surrogate models supports the effectiveness of this second phase in hybrid attacks. Although some studies \cite{wu2020making,xu2020adversarial,hu2021naturalistic,zhu2021fooling} do not explicitly describe the model theft step, they generally assume access to a surrogate model similar to one obtained via model stealing.}

\subsubsection{\textbf{gray-box attacks}}
\textcolor{black}{
This subsection covers gray-box physical adversarial attacks in computer vision, where attackers have partial knowledge that now extends from static datasets to dynamic access to the training process and detailed understanding of physical devices and their interactions. This broadens attack strategies from targeting model vulnerabilities alone to exploiting weaknesses across the entire AI system, including data pipelines, sensors, and their environment. Two main types exist: (1) data poisoning attacks that stealthily manipulate dynamic training data structures without needing full model or dataset access, undermining real-world model performance and challenging defenses focused solely on static data \cite{al2023incremental}; and (2) device-focused attacks where attackers physically or indirectly access sensors (e.g., cameras, LiDAR) to trigger adversarial effects via external signals or covertly placed physical patches. Examples include TPatch using acoustic signals to disrupt cameras \cite{zhu2023tpatch}, ITPatch exploiting rolling shutter effects with light pulses \cite{yuan2024itpatch}, AoR embedding adversarial patches in environments to mislead vSLAM \cite{chen2024adversary}, and A-HFR injecting timed spoofed LiDAR data to hide or fake objects \cite{sato2025realism}. These demonstrate a complex evolution of gray-box attacks from purely digital to physical realms.}

\textcolor{black}{These types of attacks pose significant security challenges to widely deployed Internet of Things (IoT) devices, autonomous driving systems, smart surveillance, and other physically embedded AI systems. They highlight a critical insight: protecting the digital model alone is insufficient sensors themselves and their interactions with the physical environment can also become key points of vulnerability.}

\subsection{\textbf{Simulators}}
\textcolor{black}{Although real-world evaluations represent the ultimate standard for verifying the feasibility of physical adversarial attacks in practical applications, they face several challenges, such as high cost, labor intensity, difficulty in maintaining consistent experimental conditions, and potential safety risks that make such evaluations hard to reproduce. As a result, simulated environments have become a controllable, scalable, and safe alternative for researching physical attacks.}

\textcolor{black}{Mainstream physical environment simulators are now primarily built on either Unreal Engine (UE) or Unity. UE offers high-fidelity rendering capabilities but comes with a steeper learning curve, while Unity is comparatively easier to use but provides less graphical realism. Existing scenario-based simulation engines are primarily focused on the domain of autonomous systems, such as self-driving vehicles and unmanned aerial vehicles (UAVs). These platforms provide a stable and parameterizable environment for evaluating physical attack algorithms:}

\begin{itemize}
  \item \textcolor{black}{\textbf{\emph{SVL Simulator.}} (formerly known as LGSVL \cite{rong2020lgsvl}) is an open-source autonomous driving simulation platform designed for algorithm testing. Built on the Unity3D engine, it supports integration with multiple autonomous driving stacks such as Autoware and Apollo.}
    \item \textcolor{black}{ \textbf{\emph{CARLA.}}~\cite{dosovitskiy2017carla} is an open-source autonomous driving simulator built on Unreal Engine 4. It has become the most popular simulator in the field, especially since its competitor, SVL, ceased updates in 2022. Its popularity stems from its high flexibility, strong community, comprehensive sensor support, and powerful APIs. In the autonomous vehicles field, most physical adversarial attacks are tested on this platform \cite{zhang2019camou,huang2020universal,wang2021dual,suryanto2022dta,wang2022fca,suryanto2023active}.}
\item \textcolor{black}{\textbf{\emph{AirSim.}} \cite{shah2018airsim} is an open-source simulator for autonomous systems, designed to support both self-driving vehicles and UAVs. Built on Unreal Engine 4, it focuses on autonomous visual perception and drone control, and is commonly used in academic areas, e.g., reinforcement learning and synthetic data generation.}
\end{itemize}

\subsection{\textbf{Datasets}}
\textcolor{black}{The construction of benchmarks and datasets plays a critical role in enhancing the reproducibility of physical attacks and in evaluating the robustness and stealthiness of models under realistic conditions. }

\textcolor{black}{Early studies like APRICOT \cite{braunegg2020apricot} provided a foundational resource, offering the first large, publicly available set of over 1000 "in-the-wild" images of printed adversarial patches, crucial for studying object detection model failures under diverse, real-world visual conditions. Dong et al. \cite{dong2023benchmarking} initiative established the first comprehensive benchmarks (KITTI-C, nuScenes-C, and Waymo-C) to evaluate 3D object detectors against 27 realistic corruptions in autonomous driving scenario, adopted a multi-level corruption strategy to simulate real-world environmental disruptions in camera imaging, sensoring, and LiDAR perception. Meanwhile, REAP \cite{hingun2023reap} (A Large-Scale Realistic Adversarial Patch Benchmark) offers a benchmark focusing on traffic sign recognition, providing tools to realistically render adversarial patches onto over 14,000 real-world images with differentiable transformations, allowing for scalable and authentic evaluations of patch effectiveness. Beyond mere attack success, the perceptual stealth of attacks became a key concern, addressed by the PAN \cite{li2023towards} (Physical Attack Naturalness) dataset, which uniquely incorporates human ratings and eye-tracking data from 2688 images to assess the visual naturalness of physical attacks in driving scenarios.  Notably, as one of the most comprehensive benchmarks in this domain, PADetBench \cite{lian2024padetbench} was introduced to tackle the inherent complexities and high costs associated with physical testing; this large-scale, simulation-based platform enables rigorous and fair evaluation of up to 20 attack methods against 48 different detectors by precisely controlling physical dynamics and cross-domain transformations, significantly advancing the standardization and reproducibility of physical attack research. Together, these datasets represent a multifaceted approach to deeply understanding and effectively mitigating physical adversarial threats, thereby pushing research towards developing more robust and safe AI systems.}
\begin{table*}[t]
    \centering
    \caption{\textcolor{black}{Benchmarks of physical attacks. IoU: Intersection over Union. PRC: Precision-Recall Curve. ROC: Receiver Operating Characteristic. AP: Average Precision. RCE: Relative Corruption Error. mAP: mean average precision. NDS: nuScenes detection score. mAR: mean average recall. ASR: attack success rate. SROCC: Spearman Rank Order Correlation Coefficient. PLCC: Pearson’s Linear correlation coefficient. Sc: cosine similarity. RMSE: Root Mean Squared Error. FNR: False Negative Rate.}}
    \label{tab:benchmark}
    \begin{tabularx}{\textwidth}{>{\raggedright\arraybackslash}p{3cm} >{\raggedright\arraybackslash}p{2cm} >{\raggedright\arraybackslash}p{3cm} >{\raggedright\arraybackslash}p{3.5cm} >{\raggedright\arraybackslash}p{3cm}}
        \hline
        Benchmarks & Tasks & Environments & Attacked forms & Metrics \\
        \hline
        APRICOT\cite{braunegg2020apricot} & Object Detection & Real-world & Patch & IoU, PRC, ROC \\
        \hline
        3D Corruptions \cite{dong2023benchmarking} & 3D Object Detection & Simulation & Synthesized corruptions & AP, RCE, mAP, NDS \\
        \hline
        PAN\cite{li2023towards} & Vehicle Detection & Simulation(CARLA) \& Real-world & Naturalness camouflage & SROCC, PLCC, Sc \\
        \hline
        REAP\cite{hingun2023reap} & Traffic sign recognition & Simulation & Rendering patches & RMSE, FNR, mAP, ASR \\
        \hline
        PADetBench\cite{lian2024padetbench} & Object Detection & Simulation (UE4) & Rendering patches & mAP, mAR, ASR \\
        \hline
    \end{tabularx}%
    \end{table*}

\section{\textbf{Physically Adversarial Defenses}}
\label{section3}
In this section, we review the literature on physically adversarial defenses in computer vision. It generally involves methods to mitigate or eliminate the negative effects of physically adversarial attacks. These defense processes typically occur at the image processing stage of CV systems.  Existing studies evaluate these defense mechanisms by applying digital patches directly to images or by deploying real patches in physical environments. Generally, digital adversarial patches are more effective than physical ones due to the degradation of adversarial patterns through camera imaging and image processing modules. Therefore, some studies solely conduct experiments directly on digital images, rather than first creating physically adversarial examples and capturing them with a camera (also known as physical-to-digital conversion).

Physically adversarial defense methods can be broadly divided into two categories: empirical defenses and certifiable defenses. Empirical defenses rely on the analysis of existing attacks and become effective through experiments and optimization. Certifiable defenses, on the other hand, require strict safety certification through mathematical inference or rigorous proof, ensuring the visual system's safety in specific situations. Generally, empirical defenses are more flexible and easier to implement but lack guaranteed effectiveness, requiring extensive experimentation to validate their performance. Certifiable defenses provide theoretical safety assurances but involve complex inference processes and demand more computational resources.

Table \ref{tab:Defense} provides an overview of physically adversarial defense methods from 2018 to 2024. To help readers gain a comprehensive understanding, we not only summarize the existing work into empirical defenses or certifiable defenses but also organize them according to the three stages of the computer vision system image processing procedure: pre-processing (\ref{Pre-processing}), in-processing (\ref{In-processing}), and post-processing (\ref{Post-processing}). These methods mainly focus on adversarial patch attacks. Although adversarial optical attacks and 3D-printed adversarial objects are also significant forms of attack, existing research on physical attack defenses makes it difficult to systematically investigate these two forms. Furthermore, we list the methods of patch attacks related to defense methods, along with their tasks and evaluation environments.

% %后处理
% \begin{figure}[t]
% \centering
% \includegraphics[width=0.8\textwidth]{defenses.pdf}  
% \caption{A framework for adversarial defense methods against physically adversarial attacks in different processing stages.}
% \label{Defense}
% \end{figure}

\begin{table*}[h!]
  \centering
  \caption{Representative physical defense methods against \textbf{adversarial patch attacks}. IC: Image Classification, TSR: Traffic Sign Recognition, FR: Face Recognition, PD: Person Detection, OD: Object Detection, RS: Remote Sensing, TLR: Traffic Light Recognition, VR: Video Recognition, SS: Semantic Segmentation. E: Empirical defense, C: Certifiable defense. D: Digital, P: Physical. ACM TSEM: ACM Transactions on Software Engineering and Methodology. US: USENIX Security.}
    \begin{tabular}{p{1cm}p{3cm}p{2.5cm}ccccc}
    \toprule
    Stages & Defenses   & Attacks & Tasks & Modes  & Environments & Sources\\
    \midrule
    \multirow{7}[2]{*}{Pre-} 
      & DW \cite{hayes2018visible} & \cite{brown2017adversarial}, \cite{karmon2018lavan} & IC & E & D & CVPR 2018\\
      & LGS \cite{naseer2019local} & \cite{brown2017adversarial}, \cite{karmon2018lavan} & IC & E & D & WACV 2019\\
      & SentiNet \cite{chou2020sentinet} & \cite{brown2017adversarial} & IC, TSR & E & D\&P & SPW 2020\\
      & MR \cite{mccoyd2020minority} & \cite{brown2017adversarial} & IC & E & D & ACNS 2020\\
      & BlurNet \cite{raju2020blurnet} & \cite{eykholt2018robust} & IC, TSR & E & D\&P & DSN-W 2020\\
      & TaintRadar \cite{li2021detecting} & \cite{brown2017adversarial}, \cite{sharif2016accessorize} & IC, FR & E & D\&P & INFOCOM 2021\\
      & UDF \cite{yu2022defending} & \cite{thys2019fooling}, \cite{hu2021naturalistic}, \cite{wu2020making}, \cite{xu2020adversarial} & PD & E & D\&P & TIP 2022\\
      & SAC \cite{liu2022segment} & \cite{liudpatch} & OD, TSR, RS & E & D & CVPR 2022\\
      & Jedi \cite{tarchoun2023jedi} & \cite{hu2021naturalistic} & PD & E & D\&P & CVPR 2023\\
      & PAD \cite{jing2024pad} & \cite{liudpatch}, \cite{hu2021naturalistic}, \cite{thys2019fooling} & PD & E & D\&P & CVPR 2024\\
      & DIFFender \cite{kang2023diffender} & \cite{brown2017adversarial}, \cite{karmon2018lavan}, \cite{wei2022adversarial} & IC, FR & E & D\&P & ECCV 2024\\
    \midrule
    \multirow{16}[2]{*}{In-} 
      & ROA \cite{wu2020defending} & \cite{raju2020blurnet}, \cite{sharif2016accessorize} & IC, FR, TSR & E & D & ICLR 2020\\
      & AT-LO \cite{rao2020adversarial} & \cite{brown2017adversarial}, \cite{sharif2016accessorize} & IC, TSR & E & D & ECCV 2020\\
      & IBP \cite{chiang2020certified} & \cite{karmon2018lavan} & IC & C & D & ICLR 2020\\
      & Clipped BagNet \cite{zhang2020clipped} & \cite{karmon2018lavan} & IC & C & D & SPW 2020\\
      & DeRS \cite{levine2020randomized} & \cite{karmon2018lavan} & IC & C & D & NeurIPS 2020\\
      & Ad-YOLO \cite{ji2021adversarial} & \cite{thys2019fooling} & PD & E & D\&P & Arxiv 2021\\
      & MAT \cite{metzen2021meta} & \cite{karmon2018lavan} & TLR & E & D & ICML-W 2021\\
      & RSA \cite{mu2021defending} & \cite{brown2017adversarial} & IC & C & D & ICML 2021\\
      & PatchGuard \cite{xiang2021patchguard} & \cite{karmon2018lavan} & IC & C & D & ICLR 2021\\
      & PatchCensor \cite{huang2023patchcensor} & \cite{karmon2018lavan} & IC & C & D & ACM TSEM 2023\\
      & BAGCERT \cite{metzen2021efficient} & \cite{karmon2018lavan} & IC & C & D & ICLR 2021\\
      & ScaleCert \cite{NEURIPS2021_ed519c02} & \cite{karmon2018lavan} & IC & C & D & NeurIPS 2021\\
      & MultiBN \cite{lo2021defending} & \cite{karmon2018lavan} & VR & E & D & TIP 2021\\
      & APE \cite{kim2022defending} & \cite{karmon2018lavan}, \cite{thys2019fooling}, \cite{hu2021naturalistic}, \cite{wu2020making}, \cite{xu2020adversarial} & PD & E & D\&P & ACM MM 2022\\
      & DeRS-ViT (Chen) \cite{chen2022towards} & \cite{karmon2018lavan} & IC & C & D & CVPR 2022\\
      & DeRS-ViT (Salman) \cite{salman2022certified} & \cite{karmon2018lavan} & IC & C & D & CVPR 2022\\
      & Demasked Smoothing \cite{yatsuracertified} & \cite{karmon2018lavan} & SS & C & D & ICLR 2023\\
      & PATCHCURE \cite{xiang2023patchcure} & \cite{karmon2018lavan} & IC, TSR & C & D & US 2024\\
      & NAPGuard \cite{wu2024napguard} & \cite{huang2023t}, \cite{brown2017adversarial}, \cite{wu2020making}, \cite{xu2020adversarial}, \cite{hu2022adversarial} & PD & E & D & CVPR 2024\\
    \midrule
    Post- 
      & KEMLP \cite{gurel2021knowledge} & \cite{eykholt2018robust} & TSR & E & D & ICML 2021\\
    \bottomrule
    \end{tabular}%
  \label{tab:Defense}%
\end{table*}
  
\subsection{\textbf{Pre-processing}}\label{Pre-processing}
\textcolor{black}{
Data pre-processing can mitigate the impact of physical perturbations. Current pre-processing methods can be categorized as: Adversarial Detection and completion, and image smoothing.
}

\subsubsection{\textbf{Adversarial Detection and Completion}}
\textcolor{black}{
Image completion defenses \cite{dziugaite2016study, guo2018countering, liu2022segment} first identify the patch regions by detecting differences between adversarial patches and their surrounding areas via adversarial detection. Subsequently, defenders can reconstruct these regions based on the localized information. This approach helps to neutralize the effects of adversarial attacks.
}

\textcolor{black}{
The detection of adversarial examples has garnered significant attention in recent years, primarily due to the need to mitigate the safety threats posed by these malicious inputs. A substantial body of research \cite{gong2023adversarial,zhang2018detecting,carrara2018adversarial,cohen2020detecting,xu2017feature} has focused on developing methods to distinguish adversarial examples from benign ones by leveraging auxiliary classifiers that operate on statistical features.  More recently, researchers have sought to enhance detection capabilities by exploiting neighboring statistics \cite{lee2018simple,ma2018characterizing,pang2018towards,feinman2017detecting,deng2021libre,abusnaina2021adversarial}.
}

\textcolor{black}{
Based on the above adversarial detection method in the digital world, some studies focus on detecting adversarial examples and further conduct image completion in the physical world.} 
Digital watermarking (DW) \cite{hayes2018visible} utilizes the magnitude of saliency maps to detect abnormal regions and then masks them out from the input. Unlike directly using saliency maps to locate patch regions, Jedi \cite{tarchoun2023jedi} approaches the patch localization problem from an information theory perspective. Given that the entropy of typical adversarial patches is also very high, entropy analysis can enhance the identification of potential patch regions. To better capture the shape of the patches, Jing et al. proposed PAD (Patch-Agnostic Defense) \cite{jing2024pad}. This method generates heatmaps based on the semantic and spatial heterogeneity of the patches and aligns these with masks produced by SAM models to achieve more accurate patch boundaries, thereby further improving defense effectiveness. \textcolor{black}{\cite{yu2021defending} involves gradient clipping on images to diminish informative class evidence, based on an understanding of network structure. Patch-based occlusion-aware detection (POD) \cite{strack2024defending} augments training with random patches.} 

SentiNet \cite{chou2020sentinet} employs Gradient-weighted Class Activation Mapping (Grad-CAM) to generate the mask. To mask the patch area more effectively, the SAC \cite{liu2022segment} and MR \cite{mccoyd2020minority} algorithms use segmentation techniques. The patch segmenter predicts the initial irregular patch region, which is then refined by either the shape completer \cite{liu2022segment} or through 3×3 region voting \cite{mccoyd2020minority}. This process helps recover the region covered by the masks. Recently, TaintRadar \cite{li2021detecting} has identified a strong correlation between adversarial regions and class ranks. It employs negative masks against patch attacks by alternating between three forward propagations and two backward propagations. The changes in the rank of Top-K logits observed during forward propagation guide the creation of the negative mask during back-propagation, with K logits corresponding to the K negative masks. 

For digital perturbations, researchers have concentrated on de-noising defenses against invisible noise by training image compression models \cite{jia2019comdefend} or diffusion models \cite{nie2022diffusion}. For adversarial patches, DIFFender \cite{kang2023diffender} is the first to use a diffusion model to defend against this potential risk. The approach is designed as a two-stage patch defense framework: patch localization and recovery. In the localization phase, Kang et al. \cite{kang2023diffender} found that the diffusion model exhibits a difference in its ability to recover normal areas versus adversarial patch areas in images with added Gaussian noise. This difference allows for effective identification of adversarial patches. In the recovery phase, DIFFender uses a text-guided diffusion model to remove adversarial regions from the image while preserving the integrity of the visual content. \textcolor{black}{Based on DIFender, Wei et al. \cite{wei2024real} further extend it in infrared Scenarios to defend against infrared adversarial patches.}

\subsubsection{\textbf{Image smoothing}}

\begin{figure}[ht]
\centering
\includegraphics[width=0.8\textwidth]{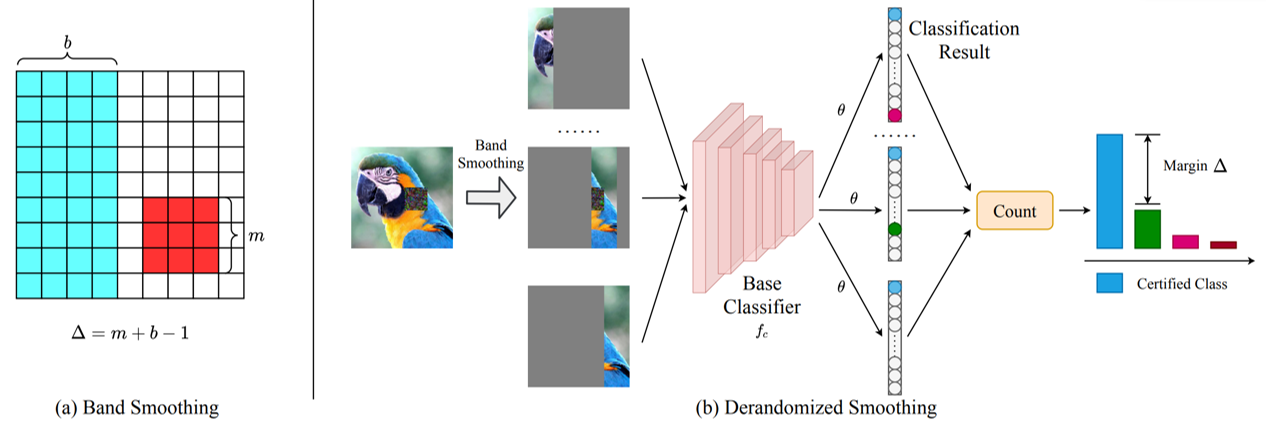} 
\caption{Introduction of the (De)Randomized Smoothing. On the left, the patch's width is $m$, and the width of the column ablation is $b$. Specifically, let $\Delta$ be the number of smoothed images affected by adversarial patches. On the right, the smoothed images are fed into the base classifier to vote for the final result. This figure comes from \cite{chen2022towards}.}
\label{DS}
\end{figure}

Randomized Smoothing \cite{cohen2019certified} involves adding Gaussian noise with varying magnitudes to an adversarial example. These noisy adversarial examples are then classified using a base classifier, and the final result is determined by voting on these classifications. (De)Randomized Smoothing (DeRS) \cite{levine2020randomized} generates smoothed images through fixed-width image ablation (illustrated in Fig.~\ref{DS}). This method allows for the calculation of how many smoothed images are affected by the adversarial patch. This quantitative framework aids in enhancing model robustness. The ViT model also employs this framework to defend against patch attacks \cite{chen2022towards, salman2022certified}. Local Gradients Smoothing (LGS) \cite{naseer2019local} directly smooths the patch regions by dividing the image into 5x5 patches and using sliding windows to identify the highest activation regions. These regions are then smoothed to remove adversarial patches. However, previous image smoothing strategies cannot ensure the integrity of the image, making them unsuitable for image segmentation tasks. Demasked Smoothing \cite{yatsuracertified} reconstructs the ablated image using ZITS \cite{dong2022incremental}, allowing the segmentation model to produce the same predictions on adversarial examples as on clean images. 

Unlike previous smoothing defenses, Yu et al. \cite{yu2022defending} uses a pre-trained Universal Defensive Noise Frame (UDF) to neutralize the impact of adversarial patches. Additionally, in the frequency domain of the image, \cite{raju2020blurnet} has found that adversarial attacks typically introduce high-frequency noise. To address this, they proposed the BlurNet model, which incorporates low-pass filtering into the feature maps of the network's first layer. This low-pass filtering effectively removes high-frequency noise, providing a defense against the RP2 attack.

\subsection{\textbf{In-processing}}\label{In-processing}

Pre-processing defenses are often time-consuming. Therefore, improving the robustness of the DNN model itself is essential. Existing methods for enhancing model robustness primarily focus on three areas: adversarial training, architecture modification, and certified robustness.

% \begin{figure}[ht]
 
% \centering
% \includegraphics[width=0.6\textwidth]{MAT.pdf}       
% \caption{MAT model trained on the Bosch Small Traffic Light datasets. This figure comes from \cite{metzen2021meta}.}
% \label{MAT}
 
% \end{figure}

\subsubsection{\textbf{Adversarial training}}
% 对抗训练的核心是将对抗样本作为训练数据来提升模型的鲁棒性
Adversarial training (AT) uses adversarial examples as training data to improve the robustness of the model \cite{madry2018towards, goodfellow2014explaining, carlini2017towards}. The goal of AT is to recover the clean prediction for adversarial examples. Researchers use physically adversarial examples to train target DNN models. Wu et al. \cite{wu2020defending} employed adversarial training (AT) to defend against rectangular occlusion attacks (ROA), which target traffic signs. Rao et al. \cite{rao2020adversarial} found that using fixed locations for adversarial patches is insufficient against location-optimized patches. Therefore, they adopted both full location optimization (covering four directions) and random location optimization (covering random directions) to enhance patch location optimization. Additionally, a model trained on a single category of patches is often deficient \cite{metzen2021meta}. Thus, patch content needs optimization as well. Meta-adversarial training (MAT) \cite{metzen2021meta} combines adversarial training with meta-learning to improve the generalization of the adversarial training model. It uses different meta-patches with varying step sizes to generate adversarial examples from an original image. MAT can achieve correct predictions for various adversarial examples. Adversarial training with patches incurs high computational overhead for large models and yields poor training results for small models, which limits its large-scale application in real-world scenarios. To address this, Zhao et al. \cite{zhao2023mitigating} employed multi-teacher adversarial robust distillation (MTARD) to guide the adversarial training of small models, balancing computational overhead with model robustness. 

\textcolor{black}{ Adversarial training with patches incurs high computational overhead for large models and yields poor training results for small models, which limits its large-scale application in real-world scenarios. To further reduce the computational overhead, a variant of adversarial training named Adversarial Robust Distillation (ARD) is proposed \cite{goldblum2020adversarially,zhu2021reliable,zi2021revisiting,zhao2022enhanced,zhao2023mitigating} for light-weight inference, which provides a potential way to make adversarial training more applicable to the physical world. Goldblum et al. \cite{goldblum2020adversarially} first introduced Adversarial Robust Distillation, which posits that employing a robust teacher model within the adversarial training framework can significantly enhance the robustness of the student model. Subsequently, Zhu et al. \cite{zhu2021reliable} proposed Incomplete Adversarial Distillation (IAD), a method that incorporates adversarial knowledge distillation by combining unreliable teacher guidance with student introspection. This approach aims to mitigate the potential negative influence of an imperfect teacher model while fostering the student model's ability to learn from its own errors. More recently, Zi et al. \cite{zi2021revisiting} presented Robust Soft Label Adversarial Distillation (RSLAD). This method leverages soft labels generated by a robust teacher model to create adversarial examples and further utilizes these robust labels to guide the training process of both clean and adversarial examples.  Zhao et al. \cite{zhao2022enhanced,zhao2023mitigating} employed multi-teacher adversarial robust distillation (MTARD) to guide the adversarial training of small models, balancing computational overhead with model robustness.}

Despite these advancements, such methods often fail to capture deep-seated characteristics of adversarial patches, such as aggressiveness and naturalness, leading to suboptimal precision and generalization against naturalistic adversarial patches (NAPs). To address this issue, NAPGuard \cite{wu2024napguard} utilizes a sophisticated critical feature modulation framework. It enhances detection capabilities against NAPs by aligning aggressive features during training and suppressing natural features during inference.

\subsubsection{\textbf{Architecture modification}}

Researchers can modify the network's architecture to mitigate the impact of adversarial patches on final classification. Existing approaches primarily focus on two aspects: 1) Narrowing the receptive field, and 2) Introducing new neural units.

\textbf{\emph{(1) Narrowing Receptive Fields.}} is inspired by BagNet \cite{brendel2018approximating}, which modifies ResNet-50 by replacing 3×3 convolutions with 1×1 convolutions and processes small image patches independently. This limits adversarial patches to affecting only a few regions. Clipped BagNet \cite{zhang2020clipped} and PatchGuard \cite{xiang2021patchguard} are built on BagNet, enhancing robustness by filtering abnormal logits or masking suspicious features. Similarly, the self-attention mechanism in ViT limits adversarial patches to a subset of tokens. PatchCensor \cite{huang2023patchcensor} uses a ViT with 16×16 patches to vote on predictions from different patches. However, narrowing receptive fields can reduce the information each feature receives, hurting model capacity and performance. PATCHCURE \cite{xiang2023patchcure} addresses this by balancing layers with small and large receptive fields, achieving effective defense with minimal performance loss.

%is inspired by BagNet \cite{brendel2018approximating}, which builds upon the ResNet-50 architecture. BagNet reduces the receptive field size by substituting the 3×3 convolutional kernels with 1×1 convolutional kernels. Additionally, BagNet processes each small image patch independently to generate per-patch logits, which are then averaged. This approach limits adversarial patches to only a few regions when attacking BagNet. Clipped BagNet \cite{zhang2020clipped} and PatchGuard \cite{xiang2021patchguard} both utilize BagNet as their backbone network. To enhance model robustness, Clipped BagNet discards abnormal logits before averaging, while PatchGuard employs a robust masking aggregation technique to detect and mask abnormal features. 

% The self-attention mechanism of ViT also constrains the "receptive field," limiting adversarial patches to affecting only a subset of the tokens. PatchCensor \cite{huang2023patchcensor} utilizes a ViT model with a 16×16 patch size \cite{dosovitskiy2020image} to perform voting on predictions from different patches. However, defense methods that narrow the receptive fields can restrict the information available to each feature, thereby reducing the capacity and performance of the DNN model. To address this, PATCHCURE \cite{xiang2023patchcure} employs a flexible model architecture that balances the number of layers with small and large receptive fields. This approach enables effective defense against adversarial patches while minimizing the impact on the model's overall performance.

\textbf{\emph{(2) Introducing New Neural Units.}} can effectively address optimization challenges when modifying a model's architecture. Ad-YOLO \cite{ji2021adversarial} adds a "Patch" class to the final layer of a YOLO-v2 network while keeping the other layers unchanged. This addition allows Ad-YOLO to both identify adversarial patches and maintain correct predictions. MultiBN \cite{lo2021defending} is a dynamic batch normalization (BN) layer structure for video recognition, addressing various types of adversarial videos, including physically adversarial ones. For each specific adversarial video, a learnable BN layer module selects the appropriate BN layer, providing more accurate distribution estimates and improving the model's generalization. AW-Net \cite{wei2024revisiting} dynamically adjusts network weights based on regulation signals from an adversarial detector, which is influenced by the input sample. This dynamic adjustment aims to enhance both accuracy and robustness. Han et al. proposed ScaleCert \cite{NEURIPS2021_ed519c02}, a robust model that focuses on superficial important neurons (neurons vulnerable to disturbances). ScaleCert masks these neurons during inference to mitigate the effects of patches. To bolster ViT's robustness, RSA \cite{mu2021defending} uses a robust aggregation (RAG) mechanism to score all tokens and discard outliers, thus reducing the impact of patch attacks. Metzen et al. introduced the BAGCERT defense \cite{metzen2021efficient}, which classifies images using a region scorer (a 3-layer CNN with kernel sizes of 3, 1, and 3) and a spatial aggregator. Adversarial Patch-Feature Energy (APE) \cite{kim2022defending} is a defense mechanism that analyzes how adversarial patches alter deep features, defining this alteration as Adversarial Patch-Feature Energy. The defense consists of APE mask, which employs layer-wise Over-Energy analysis to identify adversarial regions, and APE refinement, which mitigates patch effectiveness by refining feature pixels within the APE mask.

\textcolor{black}{
Based on the above adversarial detection method in the digital world, some studies focus on detecting adversarial examples in the physical world. Digital watermarking \cite{hayes2018visible} utilizes saliency maps to pinpoint adversarial regions and employs erosion operations to eliminate small holes within these regions. Local gradient smoothing \cite{naseer2019local} applies gradient smoothing in areas with high gradient amplitudes to mitigate the high-frequency noise resulting from patch attacks. Feature normalization and clipping \cite{yu2021defending} involve gradient clipping on images to diminish informative class evidence, based on an understanding of network structure. SAC \cite{liu2022segment} defends against patch attacks by detecting and removing patches, while Jedi \cite{tarchoun2023jedi} employs entropy to detect patches and generate masks for patches. Patch-based occlusion-aware detection (POD) \cite{strack2024defending} augments training with random patches. The diffusion-based approach DIFFender \cite{kang2024diffender}  localizes and restores adversarial regions, and Wei et al. \cite{wei2024real} further extend it in infrared images. 
}

\subsubsection{\textbf{\emph{Certified robustness.}}}

The empirical defenses mentioned earlier often lack theoretical guarantees for robustness. To address this, researchers have developed certified defenses that provide formal assurances that a model remains robust under specified conditions. A model is considered certifiably robust if, within a certain input perturbation range, the output for the true class is always confidently higher than for any other class, ensuring the prediction does not change \cite{wong2018provable}.

One prominent certified defense method is based on Interval Bound Propagation (IBP), originally designed for robustness verification in digital domains \cite{gowal2019scalable}. This approach was later extended to adversarial patch attacks \cite{chiang2020certified}. The model produces a differentiable lower bound representing the minimum margin between the true class and all others. By training to increase this margin, the model’s resistance to adversarial patches improves.

The certified defense also enforces constraints on the allowed input perturbations to guarantee robustness against all valid adversarial examples within that range.

Cohen et al. \cite{cohen2019certified} used Randomized Smoothing (RS) to provide certified robustness against digital attacks. To extend RS for defending against adversarial patches, Levine et al. \cite{levine2020randomized} proposed (De)randomized smoothing, which applies smoothing only to regions outside the patch area. Theoretically, robustness is certified when the model’s confidence in the true class exceeds the confidence in any other class by a certain margin. Even if many perturbed images are misclassified, as long as the majority with smoothing still predict the true class, robustness holds. Recently, Demasked Smoothing \cite{yatsuracertified} combines smoothing with region reconstruction to better preserve input integrity, especially for segmentation tasks.

\begin{wrapfigure}{r}{0.5\textwidth} % 图放右侧，占宽度50%
  \centering
  \includegraphics[width=0.48\textwidth]{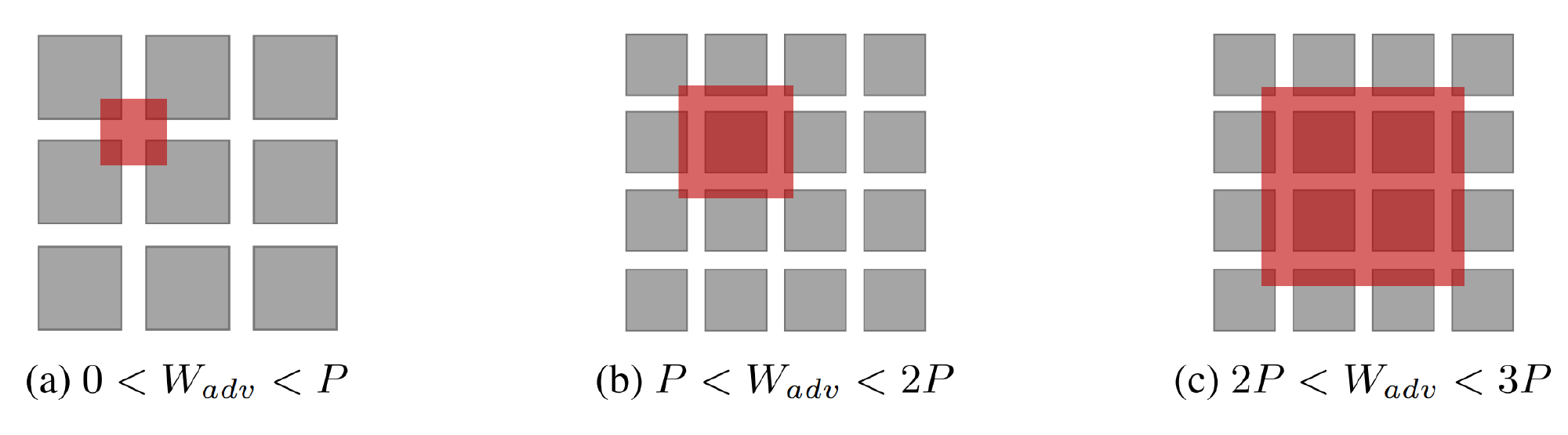}
  \caption{The number of ViT patches affected by adversarial patches with different sizes. This figure comes from \cite{huang2023patchcensor}.}
  \label{VIT}
\end{wrapfigure}
BagNet is an important model in certified defense. Clipped BagNet \cite{zhang2020clipped} achieves certified robustness by clipping abnormal feature blocks before aggregation. PatchGuard \cite{xiang2021patchguard} verifies robustness by checking the consistency of predictions under all possible maskings; consistent results certify robustness. BAGCERT \cite{metzen2021efficient} integrates training and robustness verification through a region scorer and spatial aggregator: the scorer assigns scores to image regions corresponding to true and other classes, and the aggregator calculates their difference for certification and training.

With the rise of Vision Transformers (ViT) \cite{dosovitskiy2020image}, certified defenses have begun to use ViT as the backbone. PatchCensor \cite{huang2023patchcensor} uses a pre-trained ViT model and leverages self-attention and voting across tokens to generate certified predictions. The prediction is certified if all tokens agree on the same class. As illustrated in Fig.~\ref{VIT}, at least one patch covers the adversarial patch fully, ensuring agreement among Transformer encoders. Salman et al. \cite{salman2022certified} and Chen et al. \cite{chen2022towards} combined (De)randomized smoothing with ViT by replacing global self-attention with isolated band units, significantly speeding up inference.

\subsection{\textbf{Post-processing}}
\label{Post-processing}

The mechanism of post-processing defenses aligns with the human thought process of re-verification \cite{elsayed2018adversarial}, which inspired researchers to implement such methods. After making an initial prediction, models are designed to analyze additional evidence comprehensively.

Gürel et al. \cite{gurel2021knowledge} integrated domain knowledge with adversarial defense in the Knowledge Enhanced Machine Learning Pipeline (KEMLP). KEMLP constructs factor graphs by modeling the outputs of machine learning models and predictions, where different factors have varying logical relationships with the target. For example, in the case of a "Stop" street sign, the presence of the word "Stop" serves as a sufficient condition, while the shape "Octagon" is a necessary condition. The logical assessment of each factor, combined with domain knowledge, helps form the final output, effectively preventing illogical misclassifications, such as confusing "Stop" with "speed limit."

% Inspired by the Drift-Diffusion Model (DDM) \cite{roitman2002response} and the Dropout method \cite{hinton2012improving}, Chen et al. \cite{chen2022dddm} proposed the Dropout-based Drift-Diffusion Model (DDDM). They first apply dropout to generate clipped versions of the target model. Then, the outputs of these clipped models are used for threshold accumulation. The final robust output is determined when the accumulated decision evidence reaches a specified threshold.

\section{\textbf{Outlook}}
\label{section4}
\textcolor{black}{Physically adversarial attacks and defenses have witnessed significant progress in recent years. Among them, adversarial patches have emerged as a prominent and practical threat, with demonstrable effectiveness across a variety of real-world scenarios and target models. Despite the advances, many fundamental challenges remain unresolved. In this section, we analyze the major open problems and outline prospective research directions, while highlighting their broader implications for both academic research and practical deployment.}

\textcolor{black}{(1) \textbf{\emph{Transferable} Physically Adversarial Attacks}:
In real-world scenarios such as face recognition, autonomous driving, and surveillance systems, attackers rarely have access to the model’s parameters or outputs, making white-box attacks infeasible. Therefore, the ability to craft transferable adversarial examples that generalize across different models and systems is essential. However, current transfer-based physical attacks often fail under domain shifts or variations in system architecture. This gap opens up a crucial research direction: developing more robust and generalizable transfer-based physical attacks. From a practical standpoint, understanding and improving these attacks also informs the design of more effective black-box defense systems that can anticipate such threats.}

\textcolor{black}{Future work should explore architecture-agnostic features or shared vulnerabilities across different backbone networks (e.g., ResNet vs. ViT) to guide transferable patch generation. For example, researchers can develop optimization techniques based on intermediate-layer universal perturbations or employ generative models to learn attack distributions that generalize across diverse datasets and camera conditions.}

\textcolor{black}{(2) \textbf{\emph{Robust} Physically Adversarial Attacks}:
A defining characteristic of physical attacks is their vulnerability to environmental variability, such as changes in viewing angles, distances, illumination, and camera processing. Ensuring robustness in such dynamic settings is still an unsolved problem. For researchers, this presents an opportunity to explore physics-aware optimization techniques and realistic simulation environments. From the application view, failing to account for these variabilities may lead to overestimated attack capabilities or underprepared defense mechanisms, especially in high-stakes environments like autonomous vehicles or smart surveillance.}

\textcolor{black}{Researchers should move toward 3D-aware adversarial patch generation that explicitly models transformations such as perspective warping, lighting variance, and motion blur using photorealistic renderers or differentiable simulators. Alternatively, integrating adversarial training of the patch in augmented physical simulation environments could improve robustness under real-world distortions.}

\textcolor{black}{(3) \textbf{\emph{Stealthy} Physically Adversarial Attacks}:
Beyond effectiveness, stealthiness is key for real-world adversarial examples to evade human and machine detection. Current attacks often involve conspicuous patches or patterns that are visually unnatural, making them easier to detect and filter. Future research should aim to improve the visual inconspicuousness of attacks while retaining their efficacy. Practically, enhancing stealthiness elevates the risk level of these threats, underscoring the need for advanced detection systems that can flag subtle adversarial cues without relying solely on human oversight.}

\textcolor{black}{One promising direction is to design adversarial patches that mimic natural textures (e.g., logos, road signs, or clothing patterns), leveraging semantic camouflage to blend into background scenes. Optimization constraints on patch perceptibility, e.g., using perceptual similarity metrics (LPIPS or SSIM), can further reduce visual artifacts.}

\textcolor{black}{(4) \textbf{\emph{Natural} Physically Adversarial Attacks}:
While most existing research focuses on artificially constructed adversarial examples, naturally occurring adversarial conditions—such as occlusion from fog, rain, or shadows—can unintentionally trigger misclassifications. These "natural adversaries" present a unique challenge because they cannot be anticipated or controlled in the same way as synthetic examples can. Research efforts should focus on cataloging such natural phenomena and integrating them into training and evaluation pipelines. In real-world deployment, robustness to such conditions is critical for systems operating in outdoor or uncontrolled environments, such as autonomous drones and vehicles.}

\textcolor{black}{Researchers can build benchmark datasets composed of real-world adversarial scenarios caused by natural conditions—e.g., foggy images misclassified by autonomous systems—and apply attribution methods to analyze why such failures occur. In addition, domain generalization techniques could be employed to predict misclassification risk under unseen weather or lighting conditions.}

\textcolor{black}{(5) \textbf{\emph{Universal} Physically Adversarial Defenses}:
Current physical defenses tend to be narrowly tailored to specific attack types, like adversarial patches, limiting their generalizability. The development of universal defenses that are effective against a wide range of physical perturbations is an important and open challenge. For research, this requires rethinking defense mechanisms to operate in a model-agnostic and modality-invariant manner. In practice, universal defenses can drastically reduce maintenance overhead and deployment costs, making them more viable for commercial and industrial systems.}

\textcolor{black}{Future research should explore multi-sensor fusion approaches (e.g., combining RGB with depth or thermal imaging) and modular pre-processing layers that can handle different physical perturbations. Another direction involves leveraging meta-learning to enable quick adaptation to unseen attacks.}

\textcolor{black}{(6) \textbf{\emph{Efficient} Physically Adversarial Defenses}:
Many defense mechanisms—such as adversarial training, input transformations, or sensor fusion—incur significant computational overhead, which may not be feasible for resource-constrained systems. Future work should aim to design lightweight defense strategies that balance efficacy and efficiency. For practical systems with real-time processing constraints (e.g., on-device AI in mobile or embedded platforms), the ability to defend without compromising latency or energy consumption is paramount for scalability.}

\textcolor{black}{Lightweight neural architectures with built-in robustness priors (e.g., via spatial attention masks or robust activation functions) and early-exit mechanisms can reduce inference costs. Additionally, hardware-aware defenses, such as using low-level image signal processor (ISP) filters, may mitigate attacks before they reach the neural model.}

\textcolor{black}{(7) \textbf{\emph{Trade-off} in Physically Adversarial Defenses}:
As with digital defenses, a fundamental trade-off often exists between improving robustness and preserving task accuracy. Physical defenses, in particular, must grapple with this tension under more complex and uncontrollable conditions. Investigating ways to mitigate this trade-off—through adaptive learning, ensemble models, or hybrid sensing—is a vital area for future work. Practically, understanding this balance is essential for stakeholders to make informed decisions about acceptable risk levels and system performance, especially in mission-critical applications.}

\textcolor{black}{One potential solution is task-adaptive defense tuning, where robustness is selectively prioritized for safety-critical classes (e.g., “stop sign” in autonomous driving) over less critical ones. Another is to use hybrid models that ensemble robust and high-accuracy sub-models, switching between them based on input uncertainty.}

\section{\textbf{Conclusion}}

\textcolor{black}{This survey has provided a comprehensive overview of visual adversarial attacks and defenses in the physical world, highlighting the unique challenges and emerging solutions that differ significantly from those in digital settings. We reviewed various physical attack methodologies across tasks like image classification, object detection, and face recognition, and examined how adversarial perturbations are implemented in real-world scenarios through printing, projection, and display-based strategies. On the defensive side, we categorized and analyzed robustness evaluation methods, input transformations, adversarial training approaches, and physical-world-specific defenses. Through systematic comparison, we underscored the trade-offs between robustness, computational cost, and real-world applicability. Despite progress, we identified several open challenges, including the need for standardized evaluation benchmarks, improved defense generalization, and a deeper understanding of attack-transferability in the physical domain. We hope this survey serves as a foundational reference for future research and development in building robust and secure AI vision systems in the physical world.}

\section{\textbf{Acknowledgments}}
This work was supported in part by the Project of the National Natural Science
Foundation of China under Grant 62576020 and in part by the Fundamental
Research Funds for the Central Universities.

\bibliographystyle{ACM-Reference-Format}
\bibliography{sample-base-abbre}


\end{document}